%% file: main.tex
\documentclass{article} 
\usepackage{iclr2027_conference,times}

\input{math_commands.tex}

\usepackage{hyperref}
\usepackage{url}
\usepackage{xcolor}
\usepackage{algorithm}
\usepackage{wrapfig}
\usepackage{caption}
\usepackage{amssymb}
\usepackage{graphicx} 
\usepackage{listings}
\usepackage{booktabs}

\input{inc-packages}
\input{inc-macros}

\lstdefinestyle{promptstyle}{
    basicstyle=\ttfamily\scriptsize,
    breaklines=true,
    breakatwhitespace=false,
    columns=fullflexible,
    frame=single,
    rulecolor=\color{gray!40},
    backgroundcolor=\color{gray!3},
    xleftmargin=4pt,
    xrightmargin=4pt,
    aboveskip=6pt,
    belowskip=6pt,
    showstringspaces=false
}

\algrenewcommand\algorithmicrequire{\textbf{Input:}}
\algrenewcommand\algorithmicensure{\textbf{Output:}}

\newcommand{\sminsp}{\operatorname{smin}^{\mathrm{sp}}_{\beta}}
\newcommand{\smaxsp}{\operatorname{smax}^{\mathrm{sp}}_{\beta}}

\title{Video2STL: Grounding VLM-Generated Temporal Specifications for Robot Learning}

\author{
\textbf{Merve Atasever}\textsuperscript{1} \quad
\textbf{Keyan Azbijari}\textsuperscript{1} \quad
\textbf{Cagan Bakirci}\textsuperscript{1} \quad
\textbf{Alfredo Reina Corona}\textsuperscript{1}
\\
\textbf{Bo-Ruei Huang}\textsuperscript{1} \quad
\textbf{Tolga Izdas}\textsuperscript{1} \quad
\textbf{Zahra Shahrooei}\textsuperscript{1} \quad
\textbf{Richard Yang}\textsuperscript{2}
\\
\textbf{Erdem Biyik}\textsuperscript{1} \quad
\textbf{Jyotirmoy V. Deshmukh}\textsuperscript{1}
\\[1.2ex]
\textsuperscript{1}Department of Computer Science,
University of Southern California \\
Los Angeles, CA, USA \\
\texttt{\{atasever, azbijari, cbakirci, reinacor, borueihu, izdas,}\\
\texttt{shahrooe, biyik, jdeshmuk\}@usc.edu}
\\[0.7ex]
\textsuperscript{2}University of Florida \\
Gainesville, FL, USA \\
\texttt{yangrichard@ufl.edu}
}

\iclrfinalcopy 
\begin{document}

\maketitle

\fancyhead{}

\begin{abstract}

Video-based policy learning is particularly promising, as it illustrates target behaviors without requiring action annotations or embodiment-matched demonstrations. A central challenge, however, is deciding what information should be transferred from the video to the robot. Existing approaches commonly convert visual observations into scalar similarity or value signals, or ask foundation models to directly generate reward code. While effective, these approaches can make the temporal structure of a task difficult to inspect, ground, and reuse. We present Video2STL, a framework that converts observation-only videos into parametric Signal Temporal Logic (STL) specifications and uses the resulting formal representation for robot learning. A vision-language model first extracts an embodiment-independent semantic event trace and then constructs a bank of symbolic temporal specifications. The model determines the task structure, while numerical predicate thresholds and temporal bounds are grounded from successful robot trajectories. For policy learning, we separate short- and long-timescale temporal information: short-horizon specifications provide dense rewards through rolling-window quantitative robustness, while a causal monitor over a retained long-horizon specification provides one-time progress rewards for valid temporal prefixes. The same representation is intended to support cross-embodiment transfer from human or animal videos to robot control while remaining interpretable at every stage. Across four manipulation tasks, Video2STL achieves $85.8\%$ average success-once and $67.0\%$ success-at-end, compared with $81.5\%/59.5\%$ for native dense PPO and $65.0\%/42.3\%$ for Text2Reward; in quadruped locomotion, the Qwen-3.8 and GPT-5.6-based Video2STL policies achieve $100\%$ success across velocities from $0.3$ to $2.1\,\mathrm{m/s}$ while remaining competitive in high-speed energy efficiency. Project webpage: \href{https://video2stl.github.io/}{video2stl}.


\end{abstract}

\section{Introduction}

Reinforcement learning has become a practical tool for robot control, but reward design remains one of the parts that is least reusable across tasks. A reward that works well often combines geometric errors, contact heuristics, success bonuses, regularizers, and task-specific schedules. The resulting function may train a strong policy while still being difficult to interpret or modify. This problem becomes more pronounced when the desired behavior is easier to show than to describe numerically. Human and animal videos provide abundant examples of manipulation and locomotion, yet they do not expose the demonstrator's actions in the robot action space, and the morphology of the demonstrator may be very different from that of the target robot.

Recent work has moved toward more natural interfaces for reward and behavior specification along two largely parallel lines. On the visual side, learning from human video has been framed as a cross-domain imitation problem \citep{li2022metaimitation}. VIP derives dense, goal-conditioned rewards from a representation pretrained on large-scale human video \citep{ma2023vip}, and contrastive vision-language models such as CLIP have been used directly as zero-shot, language-specified reward models \citep{rocamonde2024vlmreward}. Generative Value Learning moves further toward general progress estimation, using a VLM to infer per-frame task progress on over 300 real-world tasks spanning many robot embodiments, with human videos usable as in-context examples \citep{ma2025gvl}. On the language side, language models have served as proxy reward functions that judge behavior against a user's examples or description \citep{kwon2023rewarddesign}. Text2Reward and Eureka showed that they can also write dense reward programs and refine them through feedback or evolutionary search \citep{xie2024text2reward,ma2024eureka}. ROSETTA constructs code-based rewards from unconstrained, evolving language preferences \citep{srivastava2026rosetta}. Taken together, these results make a strong case that foundation models can provide task-level supervision. That supervision, however, takes the form of a scalar reward or progress signal for a task specified in language or as a goal image, which leaves open a question: what intermediate representation should connect an observation-only video to low-level robot learning?

A scalar reward is convenient for optimization, but it hides how the task unfolds over time. Consider a simple placement behavior. "Near the object", "move it toward the destination", "place it", and "leave it stable" are not four unrelated scores; they form a temporal structure. A reward program generated directly from text or video can encode this structure implicitly, but the structure is then mixed with numerical thresholds, simulator APIs, and implementation choices. Visual embedding rewards have the opposite advantage: they require little explicit structure, but it can be difficult to tell why a rollout receives a particular score or whether a high score corresponds to the intended temporal behavior. These limitations matter most in cross-embodiment settings, where pixel or pose correspondence is unreliable and where the transferable content is often semantic rather than kinematic.

Our central design choice is to separate \emph{what the video should decide} from \emph{what robot data should decide}. The video determines symbolic task structure: which entities matter, which semantic events occur, and how those events are ordered. Robot trajectories determine embodiment-specific numerical quantities: how close counts as \textit{Near}, how much object motion counts as \textit{Displaced}, and how long a temporal relation should be allowed to take. Signal Temporal Logic (STL) provides a natural interface between these two levels because it combines human-readable temporal operators with quantitative robustness semantics over real-valued robot signals \citep{maler2004monitoring,fainekos2009robustness,donze2010robust}. The same formula can therefore be read as a requirement, evaluated as a monitor, and converted into a dense learning signal.
\begin{figure*}[t!]
  \centering
  \includegraphics[width=\textwidth]{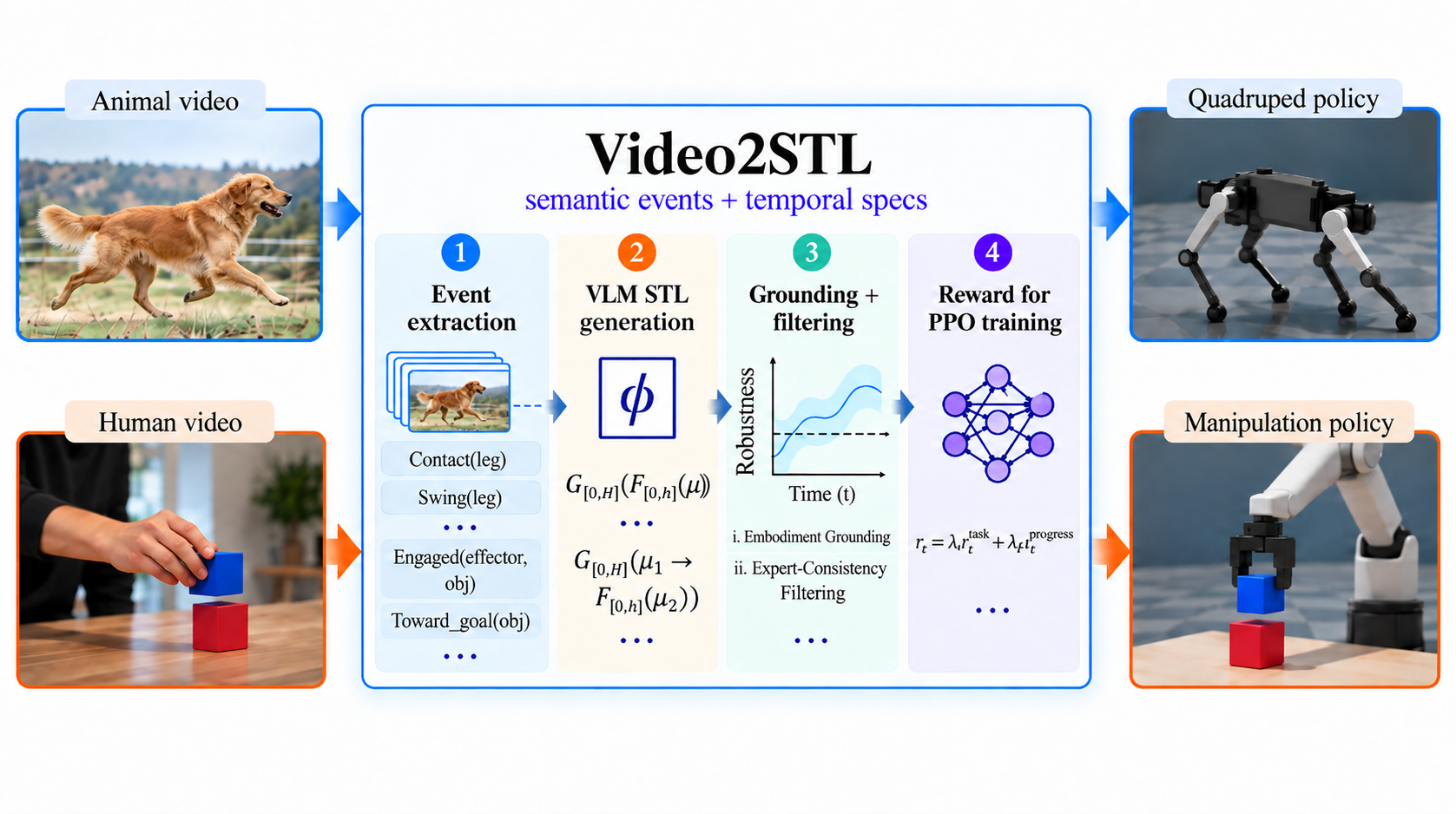}
  \vspace{-40pt}
  \caption{Overview of Video2STL framework. A VLM parses an observation video into a semantic event trace and generates parametric STL formulas. Predicate thresholds and temporal bounds are then fitted to target-embodiment trajectories. Smooth rolling-window robustness over short-horizon formulas provides dense local guidance, while a causal prefix monitor over the long-horizon backbone yields milestone progress rewards for RL policy training.}
  \label{fig:pipeline}
  \vspace{-10pt}
\end{figure*}

We introduce Video2STL, a video to formal specification pipeline for robot learning. Given an observation-only video $V$, a vision-language model (VLM) first maps the video to an embodiment-independent semantic event trace $E$. A second stage maps $E$ to a bank of parametric STL formulas $\Phi$. We deliberately do not ask the VLM to estimate robot-specific distances, velocities, or timing constants. Instead, we collect successful robot trajectories and split them once into a grounding set and a held-out filtering set. The grounding split is used to fit predicate thresholds and temporal bounds. The filtering split is used only to test whether each grounded formula is compatible with successful target-embodiment behavior.

The second design problem is how to turn a temporal specification into an RL signal without destroying credit assignment. Prior work has shown that STL can define non-Markovian rewards and encode sequential dependencies \citep{venkataraman2020tractable,puranic2021demonstrations}. An earlier locomotion work also showed that the robustness of finite-history STL can provide a structured reward once temporal templates are available \citep{atasever2026gait}. In our setting, however, the formulas are produced from video and can span very different timescales. We therefore use a two-timescale construction. Short formulas are evaluated on a trailing window $[t-H_{\mathrm{local}},t]$ and produce a dense local reward. Long formulas are not compressed into the same local window; instead, a causal prefix monitor provides a one-time reward when a new valid stage of the temporal sequence is reached. This keeps the dense signal local while preserving the ordering and deadlines of the long task.

A final issue is smooth robustness. Standard STL robustness uses nested min/max operations. Replacing them with conventional smooth approximations can improve optimization but can also change the sign of robustness, so a formula that is false under hard STL can become positive after smoothing. We use sign-preserving Boltzmann reductions whose output has the same sign as the corresponding hard min/max. This keeps the logical satisfaction boundary exact while still providing smoother within-region magnitudes for reward construction.

Our contributions are:
\begin{enumerate}
 \item We formulate observation-only video as a source of \emph{formal temporal task structure}. A two-stage VLM pipeline maps video to a semantic trace and then to a bank of parametric STL specifications, using one reusable ontology across manipulation tasks rather than task-specific reward templates.
    \item We separate symbolic inference from embodiment-specific grounding. Numerical predicate thresholds and temporal bounds are fit on one set of successful robot trajectories, while a disjoint held-out set is used for expert-consistency filtering. The expert data calibrates and validates the VLM specifications but does not supply action supervision or define their symbolic structure.
    \item We develop a two-timescale specification reward: rolling-window robustness from short formulas provides dense feedback, while a causal monitor of a long temporal backbone provides sparse, non-farmable progress. 
    \item We instantiate the framework in robot manipulation and quadruped locomotion, targeting cross-embodiment transfer in which the source demonstrator and target robot need not share morphology. The resulting representation remains inspectable from video interpretation through policy training.
\end{enumerate}

\section{Related Work}
\label{sec:related}
\myparagraph{Learning Robot Behavior from Video.}
Early work on learning from observation focused largely on closing the domain gap between human demonstrations and robot execution. This was approached through context translation between demonstrator and robot observations \citep{liu2018imitation}, viewpoint-invariant visual representations for human-to-robot imitation \citep{sermanet2018tcn}, and domain-adaptive meta-learning with paired human and robot demonstrations \citep{yu2018daml}. \citet{li2022metaimitation} further reduced the need for robot demonstrations during meta-training by translating human videos into robot-domain demonstrations. In these approaches, the information transferred from video is mainly represented through learned visual features, translated demonstrations, or adapted policies.

More recent methods use pretrained visual models to extract a learning signal directly from video. VIP learns a value-implicit representation from large-scale human video and constructs dense rewards from distances in the learned embedding space \citep{ma2023vip}. RoboCLIP compares an agent's
interaction trajectory with a video or language task specification using a pretrained video-language model and uses the resulting similarity as an episodic reward \citep{sontakke2023roboclip}. GVL instead uses a VLM to estimate task progress by reasoning over the temporal ordering of video frames, with support for in-context examples from different tasks and embodiments
\citep{ma2025gvl}.

\myparagraph{Foundation Models for Reward Design.}
A related line of work uses language and vision-language models to construct reward functions from high-level task descriptions. \citet{kwon2023rewarddesign} uses an LLM as a proxy reward function conditioned on a natural-language objective. Text2Reward generates
executable dense reward programs from language instructions and environment APIs \citep{xie2024text2reward}, while Eureka iteratively improves LLM-generated reward code using downstream policy performance as feedback
\citep{ma2024eureka}. \citet{rocamonde2024vlmreward} uses
pretrained vision-language similarity directly as an RL reward, and RL-VLM-F queries a VLM for pairwise
preferences over visual observations and trains a reward model from those comparisons \citep{wang2024rlvlmf}. Video2Reward extracts motion keypoint trajectories from the video, prompts an LLM to generate executable reward code for legged robot learning, and iteratively refines the reward using visual feedback from the learned policy \citep{zeng2024video2reward}. More recently, ROSETTA uses language preferences to construct staged reward code for manipulation \citep{srivastava2026rosetta}.

\myparagraph{Temporal Logic for Robot Learning.} Prior work has explored logic-guided RL by converting temporal-logic objectives into quantitative rewards \citep{hasanbeig2020deep, li2017reinforcement, stlrl, puranic2021demonstrations}. Other work has studied learning temporal-logic specifications from demonstrations \citep{venkataraman2020tractable, puranic2021demonstrations}. Formal specifications have also guided planning and control for legged robots, including locomotion over cluttered terrain and logic-driven gait learning for quadrupeds \cite{gu2025robust, gaitstrategiesnature, defazio2024learning, atasever2026gait}.

A related line of work translates natural-language commands into formal specifications. 
Lang2LTL grounds complex commands into LTL for long-horizon robot navigation in previously unseen environments \citep{liu2023lang2ltl}. 
The important distinction is the source and use of the specification. 
Lang2LTL starts from language and uses logic primarily to represent commands for planning/execution.
In contrast, Video2STL does not assume that the temporal task structure is given in advance: the structure is extracted from video by VLM and then checked against embodiment-specific robot data.

\begin{wrapfigure}{r}{0.5\columnwidth}
\vspace{-0.5em}
\begin{minipage}{\linewidth}

\captionsetup{type=algorithm}
\caption{\textsc{Video2STL}}
\label{alg:video2stl}

\vspace{-0.4em}
\hrule
\vspace{0.35em}

\begin{algorithmic}[1]
\setlength{\itemsep}{0.15em}

\Require Video $V$, successful trajectories $\mathcal{D}^{+}$
\Ensure Policy $\pi_\theta$

\State $E \gets \mathrm{VLM}_{\mathrm{event}}(V)$
\State $\Phi \gets \mathrm{VLM}_{\mathrm{STL}}(E)$

\State Split $\mathcal{D}^{+}
\rightarrow
(\mathcal{D}_{\mathrm{ground}},
\mathcal{D}_{\mathrm{filter}})$

\State $\vartheta \gets
\mathrm{Ground}(\Phi,\mathcal{D}_{\mathrm{ground}})$

\State $\Phi_{\mathrm{keep}}
\gets
\mathrm{Filter}(\Phi_{\vartheta},
\mathcal{D}_{\mathrm{filter}})$

\State Construct $r_t^{\mathrm{local}}$
\State Construct $r_t^{\mathrm{progress}}$

\State $r_t \gets
\lambda_L r_t^{\mathrm{local}}
+
\lambda_P r_t^{\mathrm{progress}}$

\State $\pi_\theta \gets \mathrm{RL}(r_t)$

\end{algorithmic}

\vspace{0.35em}
\hrule
\vspace{-0.3em}

\end{minipage}
\end{wrapfigure}

The closest prior lines can be summarized by what they transfer from the source signal. Video imitation methods transfer appearance or behavior representations; visual-reward methods transfer scalar similarity or progress; LLM reward-generation methods transfer language intent into executable code; language-to-logic methods transfer explicit commands into symbolic temporal constraints.
Video2STL focuses on a different interface between these directions: \emph{video $\rightarrow$ explicit temporal specification $\rightarrow$ grounded quantitative reward and causal monitor}. The proposed representation is deliberately more constrained than free-form reward code and more structured than a scalar visual reward. The temporal specifications determine which quantities must be grounded, provide explicit satisfaction boundaries that can be checked on held-out expert trajectories, and support different credit-assignment mechanisms for short- and long-horizon temporal structure.

\section{Technical Background and Preliminaries}
\label{sec:prelim}
\subsection{Problem Setting}
We consider a target robot controlled by a policy $\pi_\theta(a_t\mid o_t)$ in an episodic Markov decision process $\mathcal{M}=(\mathcal{S},\mathcal{A},P,T)$. We assume a source video $V$
showing a successful behavior, and a set of successful robot trajectories $\mathcal{D}^{+}=\{\tau^{(n)}\}_{n=1}^{N}$, where
$\tau^{(n)}=(x^{(n)}_0,\ldots,x^{(n)}_{T_n})$. The source video and robot trajectories do not need to come from the same embodiment. We first prompt the VLM to extract a semantic event trace \(E\) from the source video, identifying the task-relevant events and their temporal relationships. The VLM is then prompted a second time to convert this semantic description into a bank of parametric STL formulas \(\Phi\). At this stage, the formulas define the symbolic structure of the task but leave embodiment-specific quantities, such as geometric thresholds and temporal bounds, unspecified. Successful robot trajectories are subsequently used to ground these numerical parameters and to perform expert-consistency filtering, removing formulas that are not supported by held-out successful behavior. The method follows one principle throughout: the video determines the symbolic task structure, while robot data determines its numerical parameters. This separation allows a human or animal video to define the task without requiring morphological correspondence between the demonstrator and the robot.

\subsection{Signal Temporal Logic (STL) \& Quantitative Robustness}

Signal Temporal Logic (STL) specifies temporal properties over real-valued signals $x(t)$ using Boolean and temporal operators \citep{maler2004monitoring}.
Formulas are built from atomic predicates $f(x(t)) \ge 0$, Boolean connectives ($\wedge,\vee,\neg$), and time-bounded temporal operators $\mathbf{G}_{[a,b]}$ (always) and $\mathbf{F}_{[a,b]}$ (eventually).
We use quantitative semantics of STL \citep{donze2010robust,fainekos2009robustness}; for a given trace $x$, and each formula $\varphi$ $\rho^\varphi(t)$ assigns a real value capturing the degree of satisfaction.
Positive values imply satisfaction; negative imply violation, and the magnitude measures the satisfaction margin. 
For atomic predicates, $\rho(f(x)\ge0,x,t)=f(x(t))$.
For conjunction/disjunction, robustness uses min/max:
\(
\rho^{\varphi\wedge\psi}(t) = \min(\rho^\varphi(t),\rho^\psi(t))\), and
\(\rho^{\varphi\vee\psi}(t) = \max(\rho^\varphi(t),\rho^\psi(t)).
\)
For time-bounded operators, we have 
\(\rho^{\mathbf{G}_I\varphi}(t)=\inf_{t'\in t+I}\rho^\varphi(t')\),
and 
\(\rho^{\mathbf{F}_I\varphi}(t)=\sup_{t'\in t+I}\rho^\varphi(t')\) for an arbitrary time interval $I=[a,b]$.
Parametric Signal Temporal Logic (PSTL) extends STL by allowing constants in predicates and intervals in temporal operators to be represented by parameters \citep{asarin2011parametric}. The {\em specification mining} problem seeks to infer parameter ranges where the template formula is satisfied providing a data-driven way to instantiate temporal specifications from expert-provided locomotion trajectories.

\section{Methodology}
\label{sec:method}

\subsection{Video to an Embodiment-Independent Semantic Trace}
\label{sec:stage1}

Directly asking a VLM to write a reward from pixels entangles perception, task interpretation, robot geometry, and reward engineering in one call. We instead first ask for a semantic trace that will later be translated into temporal logic. All videos use the same global ontology for manipulation and quadruped locomotion separately. For manipulation, the current entity roles are 
\[
\mathcal{R}=\{\texttt{effector},\texttt{object},\texttt{receptacle},\texttt{handle},
\texttt{articulated\_part},\texttt{goal},\texttt{support}\},
\]
and the predicate vocabulary is
\begin{align*}
\mathcal{P}=\{&\texttt{Near}(a,b),\ \texttt{Engaged}(e,x),\ \texttt{Grasped}(o),\ 
\texttt{Released}(o),\\
&\texttt{Displaced}(o),\ \texttt{Toward\_goal}(o),\ \texttt{At\_goal}(o),\\
&\texttt{On\_support}(o,s),\ \texttt{In\_receptacle}(o,r),\ 
\texttt{Static}(x),\ \texttt{Open}(q)\}.
\end{align*}
For quadruped locomotion, the current entity roles are
\[
\mathcal{R}=\{\texttt{trunk},\texttt{front\_left},\texttt{front\_right},\texttt{hind\_left},
\texttt{hind\_right},\texttt{support}\},
\]
and the predicate vocabulary is
\begin{align*}
\mathcal{P}=\{&\texttt{Contact}(leg),\ \texttt{Swing}(leg),\ 
\texttt{Touchdown}(leg),\ 
\texttt{Liftoff}(leg),\\
&\texttt{Support\_count\_at\_least}(k,leg),\ \texttt{Trunk\_height\_stable}(t),\\ &\texttt{Trunk\_pitch\_stable}(t),\
\texttt{Trunk\_roll\_stable}(t),\ \texttt{Periodic\_leg\_motion}(leg) \}.
\end{align*}
The first query outputs a set of event records
$e_k=\big(p_k,\,\,c_k\big)$ where $p_k\in\mathcal{P}$ is the predicate and $c_k$ is an event class  (essential/terminal/incidental candidate). The output also includes pairwise temporal relations, such as \textit{before}, \textit{overlaps}, and \textit{alternates\_with}, persistent properties and a missing concepts field. If a visually important concept is absent from the ontology, the VLM must report the missing concept rather than invent a new predicate. This turns ontology insufficiency into an observable failure mode instead of silently changing the specification language.

\subsection{Semantic Trace to a Parametric STL Bank}
\label{sec:stage2}

In this stage, the VLM receives the semantic trace as a JSON file, not the original video. Its purpose is to formalize temporal relationships. The model constructs a variable-size bank $\Phi=\{\phi_1,\ldots,\phi_K\}$ using only predicates supported by the trace. Numerical predicate thresholds and time constants are forbidden at this stage, only symbolic parameters such as $H_{\mathrm{task}}$, $h_{\mathrm{goal}}$, or $h_{\mathrm{settle}}$ are allowed. The current grammar contains atomic predicates, conjunction, bounded eventually and always, and nested bounded responses: $\phi ::= \mu
\mid \phi_1\wedge\phi_2
\mid \mathbf{F}_{[0,h]}\phi
\mid \mathbf{G}_{[0,h]}\phi$ with implication represented through standard Boolean composition when needed. The nested form $\mathbf{F}_{[0,H]}
\left(
\mu_1\wedge
\mathbf{F}_{[0,h_1]}
\left(
\mu_2\wedge \mathbf{F}_{[0,h_2]}\mu_3
\right)
\right)$ expresses a bounded event sequence, while $\mathbf{F}_{[0,H]}\mathbf{G}_{[0,h]}\mu$ expresses eventual persistence. We use a bank rather than forcing a single large formula because the source video can support multiple useful requirements at different temporal scales. For example, for PushCube, the bank contains seven formulas. Representative examples are
\begin{align}
\phi_1 &= \mathbf{F}_{[0,H_{\mathrm{task}}]}
\left(
\texttt{Near}\wedge
\mathbf{F}_{[0,h_{\mathrm{engage}}]}\texttt{Engaged}
\right),
\label{eq:phi1}\\
\phi_2 &= \mathbf{F}_{[0,H_{\mathrm{task}}]}
\left(
\texttt{TowardGoal}\wedge
\mathbf{F}_{[0,h_{\mathrm{goal}}]}\texttt{AtGoal}
\right),
\label{eq:phi2}\\
\phi_5 &= \mathbf{G}_{[0,H_{\mathrm{task}}]}\texttt{OnSupport},
\label{eq:phi5}\\
\phi_7 &= \mathbf{F}_{[0,H_{\mathrm{task}}]}
\left(
\texttt{TowardGoal}\wedge
\mathbf{F}_{[0,h_{\mathrm{goal}}]}
\left(
\texttt{AtGoal}\wedge
\mathbf{F}_{[0,h_{\mathrm{settle}}]}\texttt{Static}
\right)
\right).
\label{eq:phi7}
\end{align}

The remaining formulas cover engagement-to-progress, displacement/progress, and goal-to-settling relationships. The full bank is reported in the Appendix.
\subsection{Target-Embodiment Grounding}
\label{sec:grounding}

The VLM output is symbolic. To evaluate it on a robot, each predicate must be mapped to a real-valued signed margin and each symbolic temporal parameter must be assigned a numerical bound. We split the successful trajectories into disjoint grounding and filtering sets, \(\mathcal D^+=\mathcal D_{\mathrm{ground}}\cup\mathcal D_{\mathrm{filter}}\), with \(\mathcal D_{\mathrm{ground}}\cap\mathcal D_{\mathrm{filter}}=\varnothing\). Our current experiments use a $70/30$ split. All thresholds, time bounds, and reward-normalization scales are fit from $\mathcal{D}_{\mathrm{ground}}$ only. 


Each semantic predicate $p$ is associated with a signed atomic robustness $ \rho_p(x_t;\vartheta_p)$, where $\vartheta_p$ denotes the embodiment-specific numerical parameters of the predicate and $\rho_p(x_t;\vartheta_p) \geq 0$ indicates satisfaction. We estimate $\vartheta_p$ using empirical quantiles rather than a single trajectory or hand-tuned thresholds. The choice of quantile depends on the direction of the predicate. For an upper-bound condition $p(x_t): f_p(x_t) \leq \delta_p$, we define $\rho_p(x_t)=\delta_p-f_p(x_t)$ and choose $\delta_p$ from a high quantile of the corresponding signal on successful trajectories. Conversely, for a lower-bound condition $p(x_t): f_p(x_t) \geq \delta_p$, we use $\rho_p(x_t)=f_p(x_t)-\delta_p$ and select $\delta_p$ from a suitably low quantile of successful behavior.
The use of tolerant quantiles is particularly important for predicates that appear inside temporal operators such as
$G_{[a,b]}$, whose robustness is determined by a minimum over the evaluation window. A threshold fitted to an extreme sample can make the specification unnecessarily sensitive to brief deviations. We therefore use high but non-extreme quantiles for error-type quantities, such as the $95$th percentile, and low quantiles for floor-type quantities. The same principle is applied to temporal parameters: event delays observed in successful trajectories are summarized with empirical quantiles to obtain embodiment-specific temporal bounds.

\subsection{Expert-Consistency Filtering}
\label{sec:filter}

VLM-generated formulas can be syntactically valid but incompatible with target-robot behavior. 
After grounding, we therefore evaluate each formula on the disjoint held-out set $\mathcal{D}_{\mathrm{filter}}$. 
Let $r_{ij}=\rho(\phi_i,\tau_j)$ be full-trajectory robustness of grounded formula $\phi_i$ on held-out trajectory $\tau_j$. We retain
\begin{equation}
\phi_i\in\Phi_{\mathrm{keep}}
\iff
Q_{\alpha}\!\left(\{r_{ij}:\tau_j\in\mathcal{D}_{\mathrm{filter}}\}\right)\ge 0,
\label{eq:filter}
\end{equation}
with $\alpha=0.10$ quantile in the current experiments. For PushCube task, all seven formulas pass the $Q_{0.10}\ge0$ test. Thus, this stage acts as held-out validation rather than aggressive pruning.


\subsection{Quantitative STL and Sign-Preserving Smooth Semantics}
\label{sec:robustness}


To preserve logical satisfaction while using Boltzmann-weighted
reductions, for $z\in\mathbb{R}^K$, $\beta>0$, and a nonempty
index set $A\subseteq\{1,\ldots,K\}$, define
$\operatorname{BMin}_{\beta}(z;A)
=\frac{\sum_{i\in A}e^{-\beta z_i}z_i}
{\sum_{j\in A}e^{-\beta z_j}}$;
$\operatorname{BMax}_{\beta}$ uses positive exponents.
When $\min_i z_i<0$, $\sminsp(z)$ applies
$\operatorname{BMin}_{\beta}$ only to negative coordinates;
when $\min_i z_i>0$, it uses all coordinates; otherwise it
returns zero. Symmetrically, $\smaxsp(z)$ uses positive
coordinates when $\max_i z_i>0$, all coordinates when
$\max_i z_i<0$, and returns zero otherwise. These operators preserve the satisfaction boundary:
\begin{equation}
\sminsp(z)\ge0 \iff \min_i z_i\ge0,
\qquad
\smaxsp(z)\ge0 \iff \max_i z_i\ge0.
\label{eq:signpreserve}
\end{equation}




\subsection{Two-Timescale Temporal Reward}
\label{sec:reward}

\subsubsection{Local Rolling-Window Robustness}
\label{sec:local_robustness}

Long-horizon STL specifications are useful for describing the complete task, but their robustness is not always suitable as a dense reward during policy learning. We therefore use short-horizon specifications to provide local feedback over the recent trajectory history. Let $\operatorname{span}(\phi)$ denote the temporal span required to evaluate a specification $\phi$. Given a local horizon $H_{\mathrm{local}}$, we define $\Phi_{\mathrm{local}}
=
\left\{
\phi_i \in \Phi_{\mathrm{keep}}
:
\operatorname{span}(\phi_i)
\leq H_{\mathrm{local}}
\right\}$. At time $t$, each local specification is evaluated over the trailing window $W_t
=
\left[
\max(0,t-H_{\mathrm{local}}),\,t
\right]$. 
For each $\phi_i\in\Phi_{\mathrm{local}}$, we compute its quantitative robustness $\rho^{\mathrm{sp}}_{\phi_i}(W_t)$ over $W_t$ using the sign-preserving smooth semantics. 
Because robustness magnitudes can differ substantially across formulas, we normalize each specification using a scale $s_i=Q_{\alpha}(|\rho^{\mathrm{sp}}{\phi_i}(W_t)|)$, estimated over $\tau\in\mathcal{D}{\mathrm{ground}}$, with an empirical quantile. 
The normalized robustness is
$\bar{\rho}_{\phi_i}(W_t)=
\tanh\!\left(\rho^{\mathrm{sp}}_{\phi_i}(W_t)/s_i\right)$,
which bounds each contribution to $[-1,1]$ and prevents formulas with naturally larger robustness magnitudes from dominating the reward. The local STL reward is the mean normalized robustness across the retained short-horizon specifications:
\begin{equation}
r_t^{\mathrm{local}}
=
\frac{1}{|\Phi_{\mathrm{local}}|}
\sum_{\phi_i\in\Phi_{\mathrm{local}}}
\bar{\rho}_{\phi_i}(W_t).
\label{eq:local_reward}
\end{equation}
Each formula first retains its own temporal semantics and satisfaction boundary, after which the normalized robustness values are combined to provide dense policy feedback. Specifications whose intrinsic temporal span exceeds this horizon are excluded from the local term and handled separately through the long-horizon causal progress monitor described next.






\subsubsection{Long-Horizon Causal Progress}
\label{sec:long_horizon_progress}

Short-horizon robustness provides dense feedback, but it does not by itself capture the full temporal structure of a task. A long specification may encode a sequence of events that unfolds over most of an episode, and its robustness can remain uninformative until the relevant future events occur. We therefore use a separate causal progress signal for long-horizon temporal structure.

From the retained specification bank, we select a long-horizon backbone specification $\phi_{\mathrm{backbone}} \in \Phi_{\mathrm{keep}}$ whose temporal structure represents the main progression of the task. The backbone is monitored causally: at time $t$, the monitor only uses the realized trajectory prefix $x_{0:t}$ and does not access future states. We represent the backbone as an ordered sequence of $K$ semantic stages, $p_1 \rightarrow p_2 \rightarrow \cdots \rightarrow p_K$ together with the temporal bounds inherited from the grounded STL specification. The monitor maintains the furthest valid prefix reached so far, $b_t \in \{0,1,\ldots,K\}$, where $b_t=k$ indicates that the first $k$ stages have been satisfied in the required order and within their corresponding temporal constraints.


The progress reward $r_t^{\mathrm{progress}}=(b_t-b_{t-1})/K$ is issued only when the policy reaches a new valid stage. Because $b_t$ stores the furthest valid prefix reached so far, completed stages cannot be rewarded repeatedly, and therefore $\sum_t r_t^{\mathrm{progress}}\leq1$. This construction provides sparse but temporally meaningful credit for completing valid task stages, while the rolling-window robustness term supplies dense local feedback. The final reward combines the two signals:
\begin{equation}
r_t
=
\lambda_L r_t^{\mathrm{local}}
+
\lambda_P r_t^{\mathrm{progress}}.
\label{eq:final_reward}
\end{equation}

This separation between local rolling-window robustness and long-horizon causal progress is used only for the manipulation tasks. For quadruped locomotion, we use only the local rolling-window robustness, since locomotion is primarily periodic rather than sequential and does not naturally decompose into a fixed progression of task stages.


\section{Experiments}
\label{sec:experiments}
We evaluate Video2STL in two robot-learning domains with different temporal structures: quadruped locomotion and
object manipulation. These domains provide complementary test cases for the proposed representation. Quadruped locomotion is approximately periodic with desired behavior characterized by recurring contact and stability patterns over time. Manipulation tasks, in contrast, typically consist of a sequence of semantically distinct stages, such as approaching an object, interacting with it, reaching a goal configuration, and maintaining the resulting state. Across both domains, the high-level pipeline is unchanged. Moreover, our main comparison in both domains contrasts three approaches to reward design: (i) a task-specific hand-engineered dense reward, (ii)
\textit{Text2Reward} (implemented using GPT-5.6 as
the underlying language model), which directly generates executable reward code from a natural-language task description, and (iii) \textit{Video2STL}, which uses an explicit temporal specification as the intermediate representation \citep{taomaniskill3, caluwaerts2023barkour, xie2024text2reward}.
All methods use the same policy architecture,
optimization procedure, training budget, environment configuration, and evaluation protocol; only the reward formulation is changed.

\subsection{Quadruped Locomotion}
\myparagraph{Robot and Simulation Setup.}
We evaluate locomotion on Google's Barkour vb quadruped in
MuJoCo XLA (MJX) \citep{caluwaerts2023barkour,todorov2012mujoco}. 
The control timestep is $0.02\,\mathrm{s}$ and the policy
outputs 12 normalized joint-position commands, which are converted to
actuator torques by the robot's low-level PD controller. The policy
receives the commanded linear and angular velocities together with robot proprioception and the previous action. Policies are trained using the JAX-based PPO implementation provided by Brax \citep{schulman2017proximal,freeman2021brax}.

\begin{table*}[t]
\centering
\caption{
Quadruped locomotion performance across commanded forward velocities.
Survival and success are reported as percentages, and CoT denotes cost of transportation.
Higher is better for survival and success; lower is better for CoT.
}
\label{tab:locomotion_main_results}
\resizebox{\textwidth}{!}{
\begin{tabular}{c|ccc|ccc|ccc|ccc}
\toprule
& \multicolumn{3}{c|}{\textbf{Video2STL (Qwen 3.8)}}
& \multicolumn{3}{c|}{\textbf{Video2STL (GPT-5.6)}}
& \multicolumn{3}{c|}{\textbf{Text2Reward}}
& \multicolumn{3}{c}{\textbf{Heuristic}} \\
\cmidrule(lr){2-4}
\cmidrule(lr){5-7}
\cmidrule(lr){8-10}
\cmidrule(lr){11-13}

$\mathbf{v_x}$ (m/s)
& \textbf{Surv. $\uparrow$} & \textbf{Succ. $\uparrow$} & \textbf{CoT $\downarrow$}
& \textbf{Surv. $\uparrow$} & \textbf{Succ. $\uparrow$} & \textbf{CoT $\downarrow$}
& \textbf{Surv. $\uparrow$} & \textbf{Succ. $\uparrow$} & \textbf{CoT $\downarrow$}
& \textbf{Surv. $\uparrow$} & \textbf{Succ. $\uparrow$} & \textbf{CoT $\downarrow$} \\
\midrule

0.3
& 100\% & 100\% & $2.63 \pm 0.08$
& 100\% & 100\% & $2.63 \pm 0.06$
& 100\% & 100\% & $0.91 \pm 0.01$
& 100\% & 100\% & $1.20 \pm 0.00$ \\

0.5
& 100\% & 100\% & $1.86 \pm 0.03$
& 100\% & 100\% & $1.80 \pm 0.04$
& 100\% & 100\% & $0.80 \pm 0.00$
& 100\% & 100\% & $1.00 \pm 0.00$ \\

0.7
& 100\% & 100\% & $1.58 \pm 0.02$
& 100\% & 100\% & $1.52 \pm 0.03$
& 100\% & 100\% & $0.75 \pm 0.00$
& 100\% & 100\% & $1.00 \pm 0.00$ \\

1.0
& 100\% & 100\% & $1.37 \pm 0.01$
& 100\% & 100\% & $1.26 \pm 0.01$
& 100\% & 100\% & $0.74 \pm 0.00$
& 100\% & 100\% & $1.20 \pm 0.00$ \\

1.3
& 100\% & 100\% & $1.22 \pm 0.01$
& 100\% & 100\% & $1.17 \pm 0.01$
& 100\% & 100\% & $0.79 \pm 0.00$
& 100\% & 100\% & $1.30 \pm 0.00$ \\

1.6
& 100\% & 100\% & $0.92 \pm 0.01$
& 100\% & 100\% & $1.12 \pm 0.01$
& 100\% & 100\% & $0.87 \pm 0.00$
& 100\% & 100\% & $1.40 \pm 0.00$ \\

1.9
& 100\% & 100\% & $0.98 \pm 0.01$
& 100\% & 100\% & $1.13 \pm 0.01$
& 100\% & 100\% & $1.00 \pm 0.01$
& 100\% & 100\% & $1.40 \pm 0.00$ \\

2.0
& 100\% & 100\% & $1.00 \pm 0.01$
& 100\% & 100\% & $1.13 \pm 0.01$
& 100\% & 100\% & $1.07 \pm 0.01$
& 100\% & 5\% & $1.40 \pm 0.00$ \\

2.1
& 100\% & 100\% & $1.02 \pm 0.01$
& 100\% & 100\% & $1.15 \pm 0.01$
& 100\% & 100\% & $1.14 \pm 0.01$
& 100\% & 0\% & $1.40 \pm 0.00$ \\

\bottomrule
\end{tabular}
}
\end{table*}

\myparagraph{Evaluation Metrics.}
We evaluate locomotion policies under a set of fixed commanded forward velocities with zero lateral and yaw commands. Each commanded velocity is evaluated using 20
independent rollouts of 500 simulation steps. The first 50 steps are treated as a warm-up period and excluded from command-tracking statistics. We report three task-level metrics. \emph{Survival rate} measures the fraction of rollouts that reach the full evaluation horizon without triggering an environment termination condition. \emph{Velocity-tracking success} measures whether the mean post-warm-up local forward velocity is
within $15\%$ of the commanded velocity. Specifically, for rollout $k$, $\bar{v}^{(k)}_x=(T-50)^{-1}\sum_{t=51}^{T}v^{(k)}_{x,\mathrm{loc}}(t)$ and the rollout is successful if $    \left|
    \bar{v}^{(k)}_x-v_{x,\mathrm{cmd}}
    \right|
    \leq
    0.15 |v_{x,\mathrm{cmd}}|$. We also report the cost of transportation (CoT), $\mathrm{CoT}=E/(mgd)$, where $E$ is the total energy consumed, $m$ is the robot mass, $g$ is gravitational acceleration, and $d$ is the planar distance traveled.
    

\myparagraph{Results.}
Across commanded forward velocities from $0.3$ to $2.1\,\mathrm{m/s}$, all methods achieve $100\%$ survival, indicating that survival alone does not distinguish the learned controllers. In terms of task success, Video2STL with Qwen and with GPT-5.6 maintains $100\%$ success across the entire velocity range, including the highest-speed commands. Text2Reward also achieves $100\%$ success throughout the tested range. The heuristic/hand-engineered reward baseline shows a high-speed limitation, decreasing to $5\%$ success at $2.0\,\mathrm{m/s}$ and $0\%$ at $2.1\,\mathrm{m/s}$. These results show that the Qwen- and GPT-5.6-generated specifications yield locomotion policies that remain robust across the full commanded-speed range.

The CoT reveals a complementary trade-off in locomotion efficiency. Text2Reward achieves the lowest CoT over most low- and mid-speed commands, whereas Video2STL-Qwen becomes increasingly competitive as velocity increases. In particular, at $1.9$, $2.0$, and $2.1\,\mathrm{m/s}$, Video2STL-Qwen obtains CoT values of $0.98$, $1.00$, and $1.02$, respectively, compared with $1.00$, $1.07$, and $1.14$ for Text2Reward and $1.40$ for the heuristic baseline. Video2STL-GPT-5.6 exhibits equal or lower CoT than the Qwen variant up to $1.3\,\mathrm{m/s}$ and slightly higher CoT at higher speeds, with $1.13$, $1.13$, and $1.15$ at $1.9$, $2.0$, and $2.1\,\mathrm{m/s}$. Overall, both VLM-generated specification sets yield full-range task success: Qwen provides the strongest high-speed efficiency, while Text2Reward remains more energy-efficient at lower velocities. The comparison of GPT-5.6, Gemini 3.1, and Qwen configurations is reported in Appendix~\ref{app:vlm_comparison},
Table~\ref{tab:vlm_locomotion_comparison}.

\subsection{Robot Manipulation}

\myparagraph{Tasks and Simulation Setup.}
We evaluate manipulation in ManiSkill3 \citep{taomaniskill3} using four tasks:
\textsc{PushCube-v1}, \textsc{StackCube-v1},
\textsc{LiftPegUpright-v1}, and \textsc{PlaceSphere-v1}.
Together, these tasks cover distinct forms of interaction, including
planar pushing, grasp-and-stack behavior, orientation-dependent object manipulation, and object placement. All manipulation policies use state observations and are trained with the PPO implementation provided with ManiSkill.

\myparagraph{Evaluation Metrics.}
We evaluate all methods using the native task success conditions provided by the corresponding ManiSkill environments. 
We report \emph{success-once}, the fraction of episodes in which the native success condition is reached at least once, and
\emph{success-at-end}, the fraction of episodes satisfying the native success condition in the final state. 
We evaluate all policies using the same evaluation seeds and report the mean and standard deviation across $128$ episodes.
\begin{figure*}[t!]
  \centering
  \includegraphics[width=\textwidth]{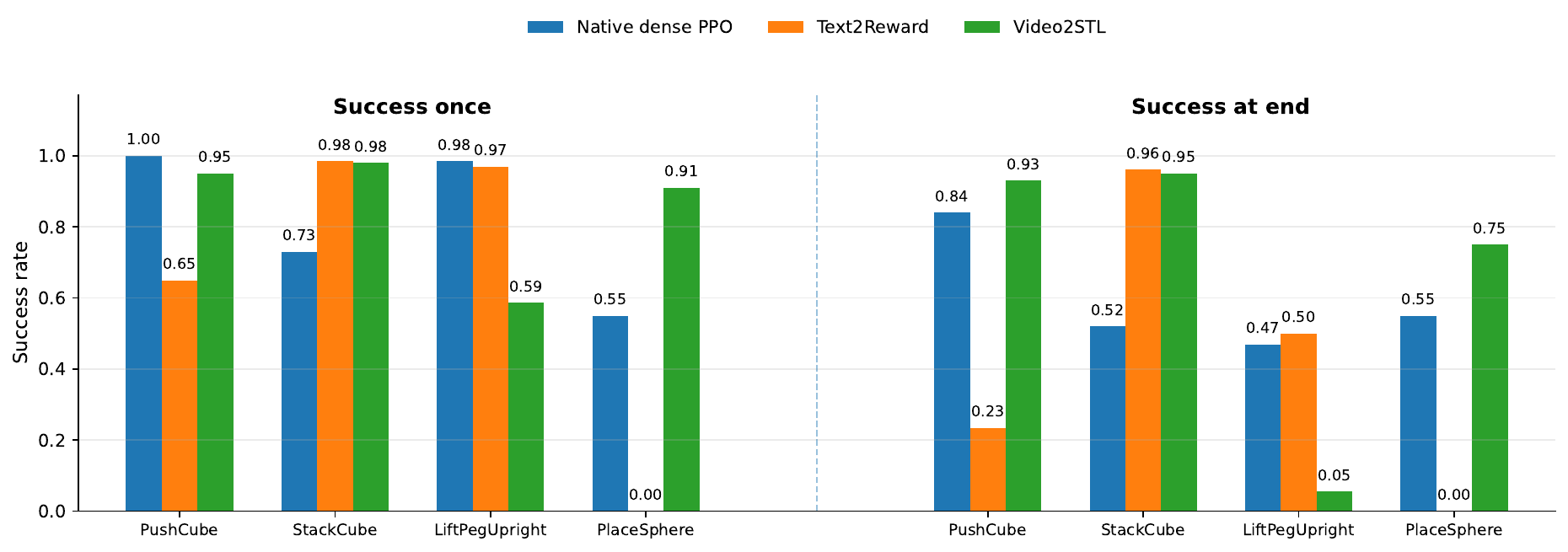}
  \caption{Manipulation results on ManiSkill3 tasks. We compare the native dense PPO reward, Text2Reward, and Video2STL on PushCube, StackCube, LiftPegUpright, and PlaceSphere.}
  \label{fig:manip_results}
  \vspace{-10pt}
\end{figure*}

\myparagraph{Results.}
Averaged over the four tasks, Video2STL reaches $85.8\%$ success-once and $67.0\%$ success-at-end, compared with $81.5\%/59.5\%$ for native dense PPO and $65.0\%/42.3\%$ for Text2Reward (Fig.~\ref{fig:manip_results}).
On PushCube, Video2STL achieves $95\%/93\%$, retaining nearly all successes.
On StackCube, it reaches $98\%/95\%$, well above native PPO ($73\%/52\%$) and on par with Text2Reward ($98\%/96\%$).
The largest gain is on PlaceSphere ($91\%/75\%$ vs.\ $55\%/55\%$ for native PPO), which Text2Reward fails to solve.
LiftPegUpright is the hardest task to retain for all methods: native PPO and Text2Reward also lose about half of their successes ($98\%/47\%$ and $97\%/50\%$).
The benchmark's check only requires the peg to be upright near the table, even while still grasped, whereas the specification extracted from the demonstration additionally requires the peg to be \emph{released} and remain static.
Video2STL thus optimizes a stricter, unassisted goal: it learns the reorientation ($59\%$ success-once) but often releases the tall, narrow peg before it settles ($5\%$ success-at-end), which we trace to a grounded \textit{Upright} tolerance ($15.9^\circ$) looser than the peg's tipping angle ($11.8^\circ$).
Overall, grounded temporal specifications yield a learning signal competitive with native dense rewards and stronger than direct reward generation on three of four tasks.

\section{Conclusion}
We presented Video2STL, a framework that converts  videos into interpretable temporal task specifications and uses them to guide robot learning across different embodiments. Across quadruped locomotion and four manipulation tasks, Video2STL demonstrated that video-derived temporal structure can provide effective learning signals without relying on manually engineered dense rewards. The results suggest that explicit temporal specifications provide a promising intermediate representation between visual demonstrations and robot learning, offering both behavioral interpretability and a structured mechanism for transferring task semantics across embodiments.

\subsection*{AI use statement}

In this work, we used generative AI tools for code generation and debugging, figure generation and refinement, manuscript drafting, and LaTeX assistance. Generative AI is also part of the proposed methodology, where VLMs extract semantic event traces from videos and generate parametric STL specifications. We did not use generative AI to fabricate experimental data or quantitative results. All AI-generated code was inspected and tested, figures were manually reviewed for technical accuracy, and all manuscript text and claims were checked by the authors. We take responsibility for the final content of this work produced with the aid of generative AI.

\bibliography{iclr2027_conference,refs}
\bibliographystyle{iclr2027_conference}

\clearpage
\appendix
\section{Appendix}

\subsection{Quadruped Locomotion}
\subsubsection{Comparison of VLM Configurations}
\label{app:vlm_comparison}

Table~\ref{tab:vlm_locomotion_comparison} reports locomotion
performance for the three VLM configurations. Their
specifications and implementation settings are detailed below.

\begin{table*}[htbp]
\centering
\small
\caption{
Quadruped locomotion performance using STL specifications generated by different VLMs.
Survival and success are reported as percentages, and CoT denotes cost of transportation.
Higher is better for survival and success; lower is better for CoT.
}
\label{tab:vlm_locomotion_comparison}
\resizebox{\textwidth}{!}{
\begin{tabular}{c|ccc|ccc|ccc}
\toprule
& \multicolumn{3}{c|}{\textbf{GPT-5.6}}
& \multicolumn{3}{c|}{\textbf{Gemini 3.1}}
& \multicolumn{3}{c}{\textbf{Qwen 3.8}} \\
\cmidrule(lr){2-4}
\cmidrule(lr){5-7}
\cmidrule(lr){8-10}

$\mathbf{v_x}$ (m/s)
& \textbf{Surv. $\uparrow$}
& \textbf{Succ. $\uparrow$}
& \textbf{CoT $\downarrow$}
& \textbf{Surv. $\uparrow$}
& \textbf{Succ. $\uparrow$}
& \textbf{CoT $\downarrow$}
& \textbf{Surv. $\uparrow$}
& \textbf{Succ. $\uparrow$}
& \textbf{CoT $\downarrow$} \\
\midrule

0.3
& 100\% & 100\% & $2.63 \pm 0.06$
& 100\% & 25\% & $2.37 \pm 0.09$
& 100\% & 100\% & $2.63 \pm 0.08$ \\

0.5
& 100\% & 100\% & $1.80 \pm 0.04$
& 100\% & 100\% & $1.97 \pm 0.04$
& 100\% & 100\% & $1.86 \pm 0.03$ \\

0.7
& 100\% & 100\% & $1.52 \pm 0.03$
& 100\% & 100\% & $1.66 \pm 0.03$
& 100\% & 100\% & $1.58 \pm 0.02$ \\

1.0
& 100\% & 100\% & $1.26 \pm 0.01$
& 100\% & 100\% & $1.26 \pm 0.02$
& 100\% & 100\% & $1.37 \pm 0.01$ \\

1.3
& 100\% & 100\% & $1.17 \pm 0.01$
& 100\% & 100\% & $1.14 \pm 0.01$
& 100\% & 100\% & $1.22 \pm 0.01$ \\

1.6
& 100\% & 100\% & $1.12 \pm 0.01$
& 100\% & 40\% & $1.16 \pm 0.01$
& 100\% & 100\% & $0.92 \pm 0.01$ \\

1.9
& 100\% & 100\% & $1.13 \pm 0.01$
& 100\% & 0\% & $1.17 \pm 0.01$
& 100\% & 100\% & $0.98 \pm 0.01$ \\

2.0
& 100\% & 100\% & $1.13 \pm 0.01$
& 95\% & 0\% & $1.17 \pm 0.01$
& 100\% & 100\% & $1.00 \pm 0.01$ \\

2.1
& 100\% & 100\% & $1.15 \pm 0.01$
& 100\% & 0\% & $1.17 \pm 0.01$
& 100\% & 100\% & $1.02 \pm 0.01$ \\

\bottomrule
\end{tabular}
}
\end{table*}

\subsubsection{Reward Construction}
For locomotion, the source observation is a video of a dog running on treadmill. (available at \href{https://www.youtube.com/watch?v=dO3G__Yqi_w}{here}) The VLM extracts embodiment-independent locomotion semantics
such as limb contact, swing, touchdown and liftoff events, support
structure, trunk stability, and periodic leg motion.

\newcommand{\stlG}{\mathbf{G}}
\newcommand{\stlF}{\mathbf{F}}
\newcommand{\stlp}[1]{\mathit{#1}}

\myparagraph{GPT-5.6.}
The semantic trace and the specification bank were produced by GPT-5.6
in a single conversation in which the video remained in context.
The bank contains 11 candidates. $\varphi_{11}$ relies on
\textit{Periodic\_leg\_motion}, a predicate that is not grounded in the
reward implementation, so it was not implemented. Each of the other ten
candidates was screened on rollouts of the trot expert (50 trajectories,
500 steps each, $v_x\in[0.5,1.6]$\,m/s), ignoring the first 50 steps of
every trajectory, and kept only if the pooled fifth percentile of its
per-step robustness was at least zero. Four candidates passed this test
($\varphi_5$, $\varphi_6$, $\varphi_9$, $\varphi_{10}$). The formulas are
listed in Table~\ref{tab:gpt_specs} and the settings in
Table~\ref{tab:gpt_parameters}.

\begin{table*}[t]
\centering
\small
\caption{
GPT-5.6 specifications and final weights. Gray formulas were
filtered out; $\varphi_{11}$ was rejected before training.
$D_\ell,L_\ell,C_\ell,S_\ell$ denote touchdown, liftoff,
contact, and swing.
$\mathcal{R}(p)=\stlG_{[0,H]}(\stlF_{[0,h_c]}p)$ and
$\mathcal{B}(p,q)=\stlG_{[0,H]}(p\rightarrow\stlF_{[0,h_s]}q)$.
$Z,P$ denote trunk-height and pitch stability.
All tanh scales are $s_i=1$.
}
\label{tab:gpt_specs}
\begin{tabular}{c|l|c}
\toprule
\textbf{ID} & \textbf{Formula} & $\mathbf{w_i}$ \\
\midrule
$\varphi_1$
& \textcolor{gray}{$\bigwedge_{\ell\in\mathcal{L}}\mathcal{R}(D_\ell)$} & 0 \\
$\varphi_2$
& \textcolor{gray}{$\bigwedge_{\ell\in\mathcal{L}}\mathcal{R}(L_\ell)$} & 0 \\
$\varphi_3$
& \textcolor{gray}{$\mathcal{R}(D_{\mathrm{FL}}\wedge D_{\mathrm{HR}})
  \wedge\mathcal{R}(D_{\mathrm{FR}}\wedge D_{\mathrm{HL}})$} & 0 \\
$\varphi_4$
& \textcolor{gray}{$\mathcal{R}(L_{\mathrm{FL}}\wedge L_{\mathrm{HR}})
  \wedge\mathcal{R}(L_{\mathrm{FR}}\wedge L_{\mathrm{HL}})$} & 0 \\
$\varphi_5$
& $\mathcal{R}(C_{\mathrm{FL}}\wedge C_{\mathrm{HR}})
  \wedge\mathcal{R}(C_{\mathrm{FR}}\wedge C_{\mathrm{HL}})$ & 1 \\
$\varphi_6$
& $\mathcal{R}(S_{\mathrm{FL}}\wedge S_{\mathrm{HR}})
  \wedge\mathcal{R}(S_{\mathrm{FR}}\wedge S_{\mathrm{HL}})$ & 1 \\
$\varphi_7$
& \textcolor{gray}{$\mathcal{B}(D_{\mathrm{FL}},D_{\mathrm{FR}})
  \wedge\mathcal{B}(D_{\mathrm{FR}},D_{\mathrm{FL}})
  \wedge\mathcal{B}(D_{\mathrm{HL}},D_{\mathrm{HR}})
  \wedge\mathcal{B}(D_{\mathrm{HR}},D_{\mathrm{HL}})$} & 0 \\
$\varphi_8$
& \textcolor{gray}{$\mathcal{B}(S_{\mathrm{FL}},S_{\mathrm{HL}})
  \wedge\mathcal{B}(S_{\mathrm{HL}},S_{\mathrm{FL}})
  \wedge\mathcal{B}(S_{\mathrm{FR}},S_{\mathrm{HR}})
  \wedge\mathcal{B}(S_{\mathrm{HR}},S_{\mathrm{FR}})$} & 0 \\
$\varphi_9$
& $\stlG_{[0,H]}Z\wedge\stlG_{[0,H]}P$ & 1 \\
$\varphi_{10}$
& $\stlG_{[0,H]}(\stlp{support\_count\_at\_least}(2))$ & 1 \\
$\varphi_{11}$
& $\bigwedge_{\ell\in\mathcal{L}}\stlG_{[0,H]}(\stlp{Periodic\_leg\_motion}(\ell))$ \ (rejected) & --- \\
\bottomrule
\end{tabular}
\end{table*}

\begin{table*}[t]
\centering
\small
\caption{
GPT-5.6 grounding and reward parameters. All values are fixed in the
reward implementation, except the support count, which is specified by
the VLM, and the filtering rule. Triples follow walk/trot/bound mode
order; $\Delta t=0.02$\,s.
}
\label{tab:gpt_parameters}
\begin{tabular}{l|l|l}
\toprule
\textbf{Parameter} & \textbf{Role} & \textbf{Value} \\
\midrule
\multicolumn{3}{l}{\emph{Temporal}} \\
$H$ & history-window setting, all modes & 30 steps \\
$h_c$ & recurrence bound & $(29,20,15)$ steps \\
$h_s$ & response bound & $(14,10,7)$ steps \\
Mode hysteresis & walk/trot entry, exit & $0.72,\ 0.65$\,m/s \\
& trot/bound entry, exit & $1.55,\ 1.45$\,m/s \\
\midrule
\multicolumn{3}{l}{\emph{Predicates}} \\
Contact / swing thresholds & foot clearance
& $0.001,\ 0.025$\,m \\
Contact / swing scales & robustness normalization
& $0.020,\ 0.025$\,m \\
Trunk-height tolerance & relative to nominal reset height
& $0.080$\,m \\
Trunk-pitch tolerance & angular bound
& $0.30$\,rad \\
$k$ & minimum support count & 2 \\
Invalid-event robustness & unavailable predecessor sample & $-8$ \\
\midrule
\multicolumn{3}{l}{\emph{Aggregation, reward, and filtering}} \\
Smoothing temperatures & atomic / temporal / formula
& $0.08,\ 0.12,\ 0.10$ \\
$s_i$ & specification tanh scale & 1.0 \\
Candidate aggregate & before warmup gate
& $\sum_i w_i\tanh(\rho_i/s_i)/\sum_i w_i$ \\
Specification weight & within total STL reward & 1.0 \\
$\tau_{\max}$ & joint torque limit & $18$\,N\,m \\
$w_{\mathrm{safe}},\alpha_{\mathrm{safe}}$
& safety weight and tanh scale & $0.25,\ 0.25$ \\
$\gamma_\tau$ & torque-effort coefficient & $10^{-6}$ \\
Reward scales & total STL / linear tracking / angular tracking
& $1.0,\ 1.0,\ 0.5$ \\
Action-rate scale & action-change penalty & $-0.01$ \\
Tracking parameter & \texttt{tracking\_sigma} & $0.25$ \\
Filter rule & retain candidate with pooled fifth percentile
& $q_{0.05}\geq0$ \\
\bottomrule
\end{tabular}
\end{table*}

\myparagraph{Qwen.}
Qwen (\texttt{qwen3.8-max-0902}) generated stages~1--2,
with the video retained in stage~2's context; GPT-5.6 generated
the reward integration. Of 15 candidates, $\varphi_{11}$ was
rejected for using disjunction. The remaining candidates were
filtered using 50 trot-expert trajectories of 500 steps,
$v_x\in[0.5,1.6]$\,m/s, discarding the first 50 steps.
A candidate was retained when the fifth percentile of its
pooled per-step monitor outputs was non-negative, leaving
seven candidates. Tables~\ref{tab:qwen_specs}
and~\ref{tab:qwen_parameters} give the formulas and settings.

\begin{table*}[t]
\centering
\small
\caption{
Qwen specifications and final weights. Gray formulas were
filtered out; $\varphi_{11}$ was rejected before training.
$D_\ell,L_\ell,C_\ell,S_\ell$ denote touchdown, liftoff,
contact, and swing.
$\mathcal{R}(p)=\stlG_{[0,H]}(\stlF_{[0,h_c]}p)$ and
$\mathcal{B}(p,q)=\stlG_{[0,H]}(p\rightarrow\stlF_{[0,h_s]}q)$.
$Z,P,R$ denote trunk-height, pitch, and roll stability.
All tanh scales are $s_i=1$.
}
\label{tab:qwen_specs}
\begin{tabular}{c|l|c}
\toprule
\textbf{ID} & \textbf{Formula} & $\mathbf{w_i}$ \\
\midrule
$\varphi_1$
& \textcolor{gray}{$\mathcal{R}(D_{\mathrm{HR}})
  \wedge\mathcal{R}(L_{\mathrm{HR}})$} & 0 \\
$\varphi_2$
& \textcolor{gray}{$\mathcal{R}(D_{\mathrm{FR}})
  \wedge\mathcal{R}(L_{\mathrm{FR}})$} & 0 \\
$\varphi_3$
& \textcolor{gray}{$\mathcal{R}(D_{\mathrm{HL}})
  \wedge\mathcal{R}(L_{\mathrm{HL}})$} & 0 \\
$\varphi_4$
& \textcolor{gray}{$\mathcal{R}(D_{\mathrm{FL}})
  \wedge\mathcal{R}(L_{\mathrm{FL}})$} & 0 \\
$\varphi_5$
& $\mathcal{B}(D_{\mathrm{HR}},D_{\mathrm{FR}})$ & 1 \\
$\varphi_6$
& $\mathcal{B}(D_{\mathrm{HL}},D_{\mathrm{FL}})$ & 0.70 \\
$\varphi_7$
& \textcolor{gray}{$\mathcal{B}(D_{\mathrm{FR}},D_{\mathrm{FL}})
  \wedge\mathcal{B}(D_{\mathrm{FL}},D_{\mathrm{FR}})$} & 0 \\
$\varphi_8$
& \textcolor{gray}{$\mathcal{B}(D_{\mathrm{HR}},D_{\mathrm{HL}})
  \wedge\mathcal{B}(D_{\mathrm{HL}},D_{\mathrm{HR}})$} & 0 \\
$\varphi_9$
& $\mathcal{R}(C_{\mathrm{HR}}\wedge C_{\mathrm{FR}})$ & 1 \\
$\varphi_{10}$
& \textcolor{gray}{$\mathcal{R}(C_{\mathrm{HL}}
  \wedge C_{\mathrm{FL}})$} & 0 \\
$\varphi_{11}$
& $\stlG_{[0,H]}C_{\mathrm{HR}}
  \vee\stlG_{[0,H]}C_{\mathrm{HR}}$ \ (rejected) & --- \\
$\varphi_{12}$
& $\stlG_{[0,H]}Z\wedge\stlG_{[0,H]}P
  \wedge\stlG_{[0,H]}R$ & 1 \\
$\varphi_{13}$
& $\stlG_{[0,H]}(\stlp{support\_count\_at\_least}(2))$ & 1 \\
$\varphi_{14}$
& $\bigwedge_{\ell\in\mathcal{L}}\mathcal{R}(C_\ell)$ & 1 \\
$\varphi_{15}$
& $\bigwedge_{\ell\in\mathcal{L}}\mathcal{R}(S_\ell)$ & 1 \\
\bottomrule
\end{tabular}
\end{table*}

\begin{table*}[t]
\centering
\small
\caption{
Qwen grounding and reward parameters, supplied by the
stage-3 configuration except the VLM-specified support
count and the filtering rule. Triples follow walk/trot/bound
mode order; $\Delta t=0.02$\,s.
}
\label{tab:qwen_parameters}
\begin{tabular}{l|l|l}
\toprule
\textbf{Parameter} & \textbf{Role} & \textbf{Value} \\
\midrule
\multicolumn{3}{l}{\emph{Temporal}} \\
$H$ & history-window setting, all modes & 30 steps \\
$h_c$ & recurrence bound & $(29,20,15)$ steps \\
$h_s$ & response bound & $(14,10,7)$ steps \\
Mode hysteresis & walk/trot entry, exit & $0.72,\ 0.65$\,m/s \\
& trot/bound entry, exit & $1.55,\ 1.45$\,m/s \\
\midrule
\multicolumn{3}{l}{\emph{Predicates}} \\
Contact / swing thresholds & foot clearance
& $0.001,\ 0.025$\,m \\
Contact / swing scales & robustness normalization
& $0.020,\ 0.025$\,m \\
Trunk-height tolerance & relative to nominal reset height
& $0.080$\,m \\
Trunk-pitch / roll tolerances & angular bounds
& $0.30,\ 0.30$\,rad \\
$k$ & minimum support count & 2 \\
Invalid-event robustness & unavailable predecessor sample & $-8$ \\
\midrule
\multicolumn{3}{l}{\emph{Aggregation, reward, and filtering}} \\
Smoothing temperatures & atomic / temporal / formula
& $0.08,\ 0.12,\ 0.10$ \\
$s_i$ & specification tanh scale & 1.0 \\
Candidate aggregate & before warmup gate
& $\sum_i w_i\tanh(\rho_i/s_i)/\sum_i w_i$ \\
Specification weight & within total STL reward & 1.0 \\
$\tau_{\max}$ & joint torque limit & $18$\,N\,m \\
$w_{\mathrm{safe}},\alpha_{\mathrm{safe}}$
& safety weight and tanh scale & $0.25,\ 0.25$ \\
$\gamma_\tau$ & torque-effort coefficient & $10^{-6}$ \\
Reward scales & total STL / linear tracking / angular tracking
& $1.0,\ 1.0,\ 0.5$ \\
Action-rate scale & action-change penalty & $-0.01$ \\
Tracking parameter & \texttt{tracking\_sigma} & $0.25$ \\
Filter rule & retain candidate with pooled fifth percentile
& $q_{0.05}\geq0$ \\
\bottomrule
\end{tabular}
\end{table*}

\myparagraph{Gemini.}
The Stage-2 output contained six candidate PSTL specifications
($\varphi_1$--$\varphi_6$). Parameters with matching counterparts in the supplied
Barkour configuration used those values; the remaining numerical groundings
retained the existing demonstration-specific settings. The candidates and the
additional torque-safety and velocity-tracking specifications were screened on
50 rollouts of the supplied expert, each lasting 500 steps with
$v_x\sim\mathcal U(0.5,1.6)$ m/s and zero lateral and yaw commands. After
discarding the first 50 steps of each rollout, a specification was retained
when the pooled fifth percentile of its smooth robustness was nonnegative.
Filtering acted on each complete specification. All six Gemini candidates,
torque safety, and forward tracking were retained; lateral and yaw tracking
were excluded.

\begin{table*}[t]
\centering
\small
\caption{
Gemini specifications and final aggregation priors. Gray formulas were
filtered out. $D_\ell$ and $L_\ell$ denote touchdown and liftoff.
$\mathcal{R}(p)=\stlG_{[0,H]}(\stlF_{[0,h_c]}p)$ and
$\mathcal{B}(p,q)=\stlG_{[0,H]}(p\rightarrow\stlF_{[0,h_p]}q)$.
$Z$ and $P$ denote trunk-height and pitch stability.
The first six rows are Gemini-generated; the remaining rows are
implementation-supplied safety and tracking terms.
}
\label{tab:gemini_specs}
\begin{tabular}{c|l|c}
\toprule
\textbf{ID} & \textbf{Formula} & $\mathbf{w_i}$ \\
\midrule
$\varphi_1$
& $\stlG_{[0,H]}Z\wedge\stlG_{[0,H]}P$ & 1 \\
$\varphi_2$
& $\stlG_{[0,H]}(\stlp{support\_count\_at\_least}(2))$ & 1 \\
$\varphi_3$
& $\mathcal{R}(D_{\mathrm{FL}})\wedge\mathcal{R}(D_{\mathrm{HR}})$ & 1 \\
$\varphi_4$
& $\mathcal{R}(L_{\mathrm{HL}})$ & 1 \\
$\varphi_5$
& $\mathcal{B}(D_{\mathrm{HL}},D_{\mathrm{FL}})$ & 1 \\
$\varphi_6$
& $\mathcal{B}(D_{\mathrm{HR}},D_{\mathrm{FR}})$ & 1 \\
\midrule
$\varphi_7$
& $\stlG_{[0,H]}\Bigl(\bigwedge_{j=1}^{12}(|\tau_j|\le\tau_{\max})\Bigr)$ & 1 \\
$\varphi_8$
& $\stlG_{[0,W]}\bigl(|v_x-v_x^{\mathrm{cmd}}|\le\epsilon_x\bigr)$ & 1 \\
track $y$
& \textcolor{gray}{$\stlG_{[0,W]}\bigl(|v_y-v_y^{\mathrm{cmd}}|\le\epsilon_y\bigr)$} & 0 \\
track yaw
& \textcolor{gray}{$\stlG_{[0,W]}\bigl(|\omega_z-\omega_z^{\mathrm{cmd}}|\le\epsilon_\omega\bigr)$} & 0 \\
\bottomrule
\end{tabular}
\par\smallskip

\end{table*}

\begin{table*}[t]
\centering
\small
\caption{
Gemini grounding and reward parameters.
Parenthesized pairs follow walk/trot mode order; unpaired values are shared;
$\Delta t=0.02$\,s.
}
\label{tab:gemini_parameters}
\begin{tabular}{l|l|l}
\toprule
\textbf{Parameter} & \textbf{Role} & \textbf{Value} \\
\midrule
\multicolumn{3}{l}{\emph{Temporal}} \\
$H$ ($H_{\mathrm{episode}}$) & outer always bound & $(30,24)$ steps \\
$h_c$ & recurrence bound & 25 steps \\
$h_p$ & phase-response bound & 13 steps \\
$W$ & velocity/yaw tracking bound & $(30,24)$ steps \\
Mode hysteresis & trot entry / walk return & $0.72,\ 0.65$\,m/s \\
\midrule
\multicolumn{3}{l}{\emph{Predicates}} \\
$z_{\mathrm{th}}$ & contact clearance threshold & $0.003$\,m \\
$c$ & contact robustness scale & $0.002$\,m \\
Contact / swing & opposite signed margins & $C_\ell$ / $-C_\ell$ \\
Touchdown / liftoff & exact contact transitions & adjacent samples \\
$z_{\mathrm{ref}}$ & reference trunk height & torso height at reset \\
$\delta_z,s_z$ & height tolerance / margin scale & $0.06$\,m / $0.02$\,m \\
$\theta_{\max}$ & absolute trunk-pitch tolerance & $(8^\circ,7^\circ)$ \\
$s_\theta$ & pitch robustness scale & $5^\circ$ \\
$k$ & support count ($k$th-largest margin) & 2 \\
$\epsilon_x$ & forward tracking tolerance & $(0.55,0.60)$\,m/s \\
$\epsilon_y,\epsilon_\omega$ & excluded lateral/yaw tolerances & $0.05$\,m/s; $0.05$\,rad/s \\
\midrule
\multicolumn{3}{l}{\emph{Aggregation, reward, and filtering}} \\
$\beta_{\mathrm{atom}},\beta_{\mathrm{temp}}$ & Boolean / temporal sharpness & $20,\ 20$ \\
$\beta_{\mathrm{group}}$ & group Boltzmann soft-min sharpness & 0.5 \\
$w_i$ & specification prior & 1 active; 0 excluded \\
$\alpha_g$ & group tanh scale & 1 \\
$w_{\mathrm{safe}},w_{\mathrm{track}}$ & safety / tracking group weights & $1,\ 1$ \\
$w_{\mathrm{pattern}}$ & gait-pattern group weight & $(1.1,1.2)$ \\
$\tau_{\max},s_\tau$ & torque limit / margin scale & $18$\,N\,m / $5$\,N\,m \\
Forward multiplier & multiplier of $\rho_x/\epsilon_x$ & 1 \\
Total reward rate & sum of bounded group rewards & $\sum_g w_g\tanh(\rho_g/\alpha_g)$ \\
Filter rule & pooled retention criterion & $q_{0.05}\ge0$ \\
\bottomrule
\end{tabular}
\par\smallskip
\end{table*}

\subsubsection{Hyperparameters and Domain Randomization}

We train each policy for $400$M environment steps using an unroll
length of 30, 32 minibatches, 4 updates per batch, discount factor
$\gamma=0.955$, learning rate $1.5\times 10^{-4}$, entropy coefficient
$0.004$, 8192 parallel environments, and a batch size of 256. All reward
variants use the same policy architecture and PPO optimization settings.


We additionally apply domain
randomization to friction and actuator parameters. Friction is sampled in
the range $(0.6,1.4)$, while actuator gain and bias perturbations are
sampled in the range $(-5,5)$.

\subsubsection{Sources Videos \& VLM Prompts For Quadruped Locomotion}

We use two sequential VLM prompts for quadruped locomotion.
The first extracts an embodiment-independent semantic locomotion trace
directly from an animal video. The second receives only
this structured trace and converts it into candidate parametric STL
specifications. 

\subsubsubsection{Stage 1: Video to Semantic Locomotion Trace}

\begin{lstlisting}[style=promptstyle]
System Role:
Act as an expert Vision-Language Model specializing in Biomechanics and Robotic Locomotion.

Context & Objective:
Analyze the provided video of a real quadruped animal performing locomotion. Your goal is to extract an embodiment-independent semantic description of the temporal locomotion structure. This analysis will later be used to construct Signal Temporal Logic (STL) specifications for a quadruped robot with different morphology, body dimensions, and controllers. Therefore, you must prioritize repeated, behaviorally important temporal structures over exact joint angles, limb trajectories, or incidental movements.

Ontology & Permitted Vocabulary:
Use only the following terms for your analysis.

1. Body Roles:

Trunk: The main body/torso of the quadruped.
Front_left: Front-left leg/foot.
Front_right: Front-right leg/foot.
Hind_left: Hind-left leg/foot.
Hind_right: Hind-right leg/foot.
Support_surface: The surface supporting locomotion, including ground or treadmill belt.

2. States & Events:

Contact(leg): Foot is visibly in contact with the support surface.
Swing(leg): Leg is in the aerial/swing phase.
Touchdown(leg): Transition from swing to contact.
Liftoff(leg): Transition from contact to swing.
Support_count_at_least(k): At least k feet are supporting the body.
Trunk_height_stable: The trunk maintains approximately bounded vertical variation over repeated gait cycles.
Trunk_pitch_stable: The trunk maintains approximately bounded pitch variation over repeated gait cycles.
Trunk_roll_stable: The trunk maintains approximately bounded roll variation over repeated gait cycles.
Periodic_leg_motion(leg): The leg exhibits a repeated locomotion cycle.

3. Temporal Relations:

before
after
overlaps
alternates_with (Events repeatedly occur in alternating portions of the gait cycle.)
approximately_synchronous_with (Events repeatedly occur close together.)
phase_shifted_from (Consistent nonzero temporal offset between two legs.)

4. Observation Classes:

Essential_candidate: Visually supported, embodiment-independent temporal relationships critical to describing the locomotion pattern.

Persistent_candidate: Body-level properties persisting through multiple gait cycles
(e.g., stable trunk height).

Incidental: Visually present but unnecessary for formal description
(e.g., head movement, tail motion, panting, a single irregular step).

Strict Constraints & Rules:

- Do NOT guess: If left/right identity or contacts cannot be reliably determined due to viewpoint or occlusion, explicitly report the uncertainty.

- Do NOT infer robot-specific metrics: Exclude joint angles, torques, actions, motor states, joint velocities, actuator properties, or numerical thresholds.

- Treadmill physics: If the animal is on a treadmill, its body may remain stationary in image/world coordinates. Do NOT interpret this as zero forward velocity.

Task Requirements:
Analyze the video and determine:

1. Visible body parts.
2. Repeated stance/contact and swing behaviors.
3. Repeated touchdown and liftoff ordering.
4. Strongest inter-leg temporal relationships
   (synchronization, alternation, phase offsets).
5. Persistent body-level regularities
   (e.g., stable trunk pitch).
6. Ambiguities and occlusions.

Output Format:

Return valid JSON only, using exactly this structure.
Do not include markdown code blocks or conversational text outside the JSON.

{
  "locomotion_context": {
    "type": "overground | treadmill | uncertain",
    "confidence": 0.0,
    "evidence": "brief visual justification"
  },

  "visibility": {
    "trunk": "clear | partial | unclear",
    "front_left": "clear | partial | unclear",
    "front_right": "clear | partial | unclear",
    "hind_left": "clear | partial | unclear",
    "hind_right": "clear | partial | unclear"
  },

  "observed_events": [
    {
      "event_id": "e1",
      "predicate": "touchdown(front_left)",
      "event_class":
        "essential_candidate | persistent_candidate | incidental",
      "confidence": 0.0,
      "visual_evidence": "brief explanation"
    }
  ],

  "repeated_states": [
    {
      "predicate": "contact(front_left)",
      "confidence": 0.0,
      "visual_evidence": "brief explanation"
    }
  ],

  "inter_leg_relations": [
    {
      "relation_id": "r1",
      "leg_1": "front_left",
      "relation":
        "before | after | overlaps | alternates_with |
         approximately_synchronous_with | phase_shifted_from",
      "leg_2": "hind_right",
      "event_basis": "touchdown | liftoff | contact | swing",
      "confidence": 0.0,
      "evidence": "brief explanation"
    }
  ],

  "persistent_properties": [
    {
      "predicate":
        "Trunk_height_stable | support_count_at_least(k)",
      "confidence": 0.0,
      "evidence": "brief explanation"
    }
  ],

  "uncertainties": [
    "brief description of an ambiguity or occlusion"
  ]
}
\end{lstlisting}

\subsubsubsection{Stage 2: Semantic Trace to PSTL Specifications}

\begin{lstlisting}[style=promptstyle]
System Role:
Act as an expert Formal Methods Engineer specializing in Logic Synthesis for Robotic Control.

Context & Objective:
You are constructing Parametric Signal Temporal Logic (PSTL) specifications based on a semantic locomotion trace. This trace (provided as a JSON input) was extracted from a video of a real quadruped animal. Your objective is to translate this trace into embodiment-independent, candidate temporal specifications that describe the demonstrated repeated locomotion pattern. These specifications will eventually be grounded to a four-legged robot with a different morphology to generate a dense, continuous reward signal for reinforcement learning.

Strict Constraints (The "Zero-Hallucination" Rule):

- Trace-Dependent Reasoning: You must reason only from the supplied semantic trace.

- No Invention: Do not invent contact events, predicates, or temporal relations that are absent from the input data. Use a predicate only if it is explicitly supported by the trace.

- Symbolic Grounding: All temporal constants and numerical temporal parameters must remain purely symbolic (e.g., H_episode, h_cycle) for later numerical grounding during training.

Allowed PSTL Grammar:

You are restricted to the following exact textual grammar.
(Note: mu represents an atomic predicate; phi represents a sub-formula; H and h represent symbolic temporal bounds).

Atomic predicate:
mu

Conjunction:
phi_1 AND phi_2

Bounded eventuality:
F_[0,H](mu)

Bounded persistence:
G_[0,H](mu)

Repeated-event requirement:
G_[0,H](F_[0,h](mu))

Response requirement:
G_[0,H](mu_1 -> F_[0,h](mu_2))

Repeated paired-event requirement:
G_[0,H](F_[0,h](mu_1 AND mu_2))

Ordered repeated response:
G_[0,H](mu_1 -> F_[0,h1](mu_2))

Note:
Conjunctions of independently valid formulas are permitted.

Output Format:

Return valid JSON only, using exactly the structure below.
Do not include markdown code blocks, conversational text, or explanations outside the JSON.

{
  "candidate_specifications": [
    {
      "candidate_id": "phi_1",

      "formula":
        "PSTL formula using ONLY the allowed textual grammar",

      "predicates_used": [
        "instantiated_predicate_1"
      ],

      "supporting_event_ids": [
        "e1"
      ],

      "supporting_relation_ids": [
        "r1"
      ],

      "excluded_incidental_event_ids": [
        "e7"
      ],

      "symbolic_parameters": [
        "H_episode",
        "h_cycle"
      ],

      "semantic_complexity": 1,

      "rationale":
        "Brief explanation grounded purely in the input semantic trace"
    }
  ],

  "stage1_ambiguities_affecting_specification": [
    "Brief description of how an ambiguity in the input trace impacts the formula"
  ]
}
\end{lstlisting}

\subsection{Robot Manipulation}

\subsubsection{Reward Construction}

This section lists the parametric Signal Temporal Logic (PSTL)
specifications generated by the VLM from the semantic traces of the
manipulation demonstrations. These are the raw candidate specifications
produced before numerical grounding using robot expert trajectories and
before expert-consistency filtering. Thus, not every specification listed
below is necessarily retained in the final reward.

For compactness, we denote the manipulated object by $o$, the robot
end-effector by $e$, the support surface by $s$, and the receptacle by $r$.
All temporal bounds, such as $H_{\mathrm{task}}$ and $h_{\mathrm{goal}}$,
are symbolic at this stage and are subsequently grounded from expert
trajectories.

\subsubsubsection{PushCube-v1}
\label{app:pushcube_specs}

For \textsc{PushCube}, the VLM generated the following eight candidate specifications:
\begin{align}
\phi_1:\quad
&\mathbf{F}_{[0,H_{\mathrm{task}}]}
\left(
\mathrm{Near}(e,o)
\land
\mathbf{F}_{[0,h_{\mathrm{engage}}]}
\mathrm{Engaged}(e,o)
\right),
\\
\phi_2:\quad
&\mathbf{F}_{[0,H_{\mathrm{task}}]}
\left(
\mathrm{Toward\_goal}(o)
\land
\mathbf{F}_{[0,h_{\mathrm{goal}}]}
\mathrm{At\_goal}(o)
\right),
\\
\phi_3:\quad
&\mathbf{F}_{[0,H_{\mathrm{task}}]}
\left(
\mathrm{Engaged}(e,o)
\land
\mathbf{F}_{[0,h_{\mathrm{progress}}]}
\left(
\mathrm{Toward\_goal}(o)
\land
\mathbf{F}_{[0,h_{\mathrm{goal}}]}
\mathrm{At\_goal}(o)
\right)
\right),
\\
\phi_4:\quad
&\mathbf{F}_{[0,H_{\mathrm{task}}]}
\mathrm{Displaced}(o)
\land
\mathbf{F}_{[0,H_{\mathrm{task}}]}
\mathrm{Toward\_goal}(o),
\\
\phi_5:\quad
&\mathbf{G}_{[0,H_{\mathrm{task}}]}
\mathrm{On\_support}(o,s),
\\
\phi_6:\quad
&\mathbf{F}_{[0,H_{\mathrm{task}}]}
\left(
\mathrm{At\_goal}(o)
\land
\mathbf{F}_{[0,h_{\mathrm{settle}}]}
\mathrm{Static}(o)
\right),
\\
\phi_7:\quad
&\mathbf{F}_{[0,H_{\mathrm{task}}]}
\left(
\mathrm{Toward\_goal}(o)
\land
\mathbf{F}_{[0,h_{\mathrm{goal}}]}
\left(
\mathrm{At\_goal}(o)
\land
\mathbf{F}_{[0,h_{\mathrm{settle}}]}
\mathrm{Static}(o)
\right)
\right),
\\
\phi_8:\quad
&\mathbf{F}_{[0,H_{\mathrm{task}}]}
\left(
\mathrm{Toward\_goal}(o)
\land
\mathbf{F}_{[0,h_{\mathrm{goal}}]}
\mathrm{At\_goal}(o)
\right)
\land
\mathbf{G}_{[0,H_{\mathrm{task}}]}
\mathrm{On\_support}(o,s).
\end{align}
These specifications capture approach and engagement, object displacement,
goal-directed progress, goal attainment, maintenance of support, and
post-goal stationarity.

\subsubsubsection{StackCube-v1}
\label{app:stackcube_specs}

For \textsc{StackCube}, the VLM generated seven candidate specifications:
\begin{align}
\phi_1:\quad
&\mathbf{F}_{[0,H_{\mathrm{task}}]}
\left(
\mathrm{Grasped}(o)
\land
\mathbf{F}_{[0,h_{\mathrm{displace}}]}
\mathrm{Displaced}(o)
\right),
\\
\phi_2:\quad
&\mathbf{F}_{[0,H_{\mathrm{task}}]}
\left(
\mathrm{Grasped}(o)
\land
\mathbf{F}_{[0,h_{\mathrm{reach}}]}
\left(
\mathrm{Toward\_goal}(o)
\land
\mathbf{F}_{[0,h_{\mathrm{goal}}]}
\mathrm{At\_goal}(o)
\right)
\right),
\\
\phi_3:\quad
&\mathbf{G}_{[0,H_{\mathrm{task}}]}
\left(
\mathrm{Toward\_goal}(o)
\rightarrow
\mathbf{F}_{[0,h_{\mathrm{goal}}]}
\mathrm{At\_goal}(o)
\right),
\\
\phi_4:\quad
&\mathbf{F}_{[0,H_{\mathrm{task}}]}
\left(
\mathrm{On\_support}(o,s)
\land
\mathbf{F}_{[0,h_{\mathrm{release}}]}
\left(
\mathrm{Released}(o)
\land
\mathbf{F}_{[0,h_{\mathrm{settle}}]}
\mathrm{Static}(o)
\right)
\right),
\\
\phi_5:\quad
&\mathbf{F}_{[0,H_{\mathrm{task}}]}
\left(
\mathbf{G}_{[0,h_{\mathrm{support}}]}
\mathrm{On\_support}(o,s)
\right),
\\
\phi_6:\quad
&\mathbf{F}_{[0,H_{\mathrm{task}}]}
\left(
\mathbf{G}_{[0,h_{\mathrm{static}}]}
\mathrm{Static}(o)
\right),
\\
\phi_7:\quad
&\mathbf{F}_{[0,H_{\mathrm{task}}]}
\left(
\mathrm{Grasped}(o)
\land
\mathbf{F}_{[0,h_{\mathrm{reach}}]}
\left(
\mathrm{Toward\_goal}(o)
\land
\mathbf{F}_{[0,h_{\mathrm{goal}}]}
\mathrm{At\_goal}(o)
\right)
\right)
\nonumber\\
&\land
\mathbf{F}_{[0,H_{\mathrm{task}}]}
\left(
\mathrm{On\_support}(o,s)
\land
\mathbf{F}_{[0,h_{\mathrm{release}}]}
\left(
\mathrm{Released}(o)
\land
\mathbf{F}_{[0,h_{\mathrm{settle}}]}
\mathrm{Static}(o)
\right)
\right)
\nonumber\\
&\land
\mathbf{F}_{[0,H_{\mathrm{task}}]}
\left(
\mathbf{G}_{[0,h_{\mathrm{support}}]}
\mathrm{On\_support}(o,s)
\right)
\land
\mathbf{F}_{[0,H_{\mathrm{task}}]}
\left(
\mathbf{G}_{[0,h_{\mathrm{static}}]}
\mathrm{Static}(o)
\right).
\end{align}

The resulting specification set describes grasping and transport toward
the target, supported placement, release, and persistent terminal
stability.

\subsubsubsection{LiftPegUpright-v1}
\label{app:liftpeg_specs}

For \textsc{LiftPegUpright}, the VLM generated seven candidate
specifications:
\begin{align}
\phi_1:\quad
&\mathbf{F}_{[0,H_{\mathrm{task}}]}
\left(
\mathrm{Near}(e,o)
\land
\mathbf{F}_{[0,h_{\mathrm{reach}}]}
\mathrm{Engaged}(e,o)
\right),
\\
\phi_2:\quad
&\mathbf{F}_{[0,H_{\mathrm{task}}]}
\left(
\mathrm{Grasped}(o)
\land
\mathbf{F}_{[0,h_{\mathrm{manip}}]}
\mathrm{Displaced}(o)
\right),
\\
\phi_3:\quad
&\mathbf{F}_{[0,H_{\mathrm{task}}]}
\left(
\mathrm{Toward\_goal}(o)
\land
\mathbf{F}_{[0,h_{\mathrm{goal}}]}
\mathrm{Upright}(o)
\right),
\\
\phi_4:\quad
&\mathbf{F}_{[0,H_{\mathrm{task}}]}
\left(
\mathrm{Upright}(o)
\land
\mathbf{F}_{[0,h_{\mathrm{terminal}}]}
\left(
\mathrm{On\_support}(o,s)
\land
\mathbf{F}_{[0,h_{\mathrm{terminal2}}]}
\mathrm{At\_goal}(o)
\right)
\right),
\\
\phi_5:\quad
&\mathbf{F}_{[0,H_{\mathrm{task}}]}
\left(
\mathrm{Upright}(o)
\land
\mathbf{F}_{[0,h_{\mathrm{release}}]}
\left(
\mathrm{Released}(o)
\land
\mathbf{F}_{[0,h_{\mathrm{stable}}]}
\mathrm{Static}(o)
\right)
\right),
\\
\phi_6:\quad
&\mathbf{F}_{[0,H_{\mathrm{task}}]}
\left(
\mathbf{G}_{[0,h_{\mathrm{persist}}]}
\mathrm{Upright}(o)
\right),
\\
\phi_7:\quad
&\mathbf{F}_{[0,H_{\mathrm{task}}]}
\left(
\mathbf{G}_{[0,h_{\mathrm{persist2}}]}
\mathrm{On\_support}(o,s)
\right).
\end{align}

These candidates characterize approach and interaction with the peg,
displacement, progress toward an upright configuration, supported
placement, release followed by stability, and persistence of the terminal
upright and supported states.

\subsubsubsection{PlaceSphere-v1}
\label{app:placesphere_specs}

For \textsc{PlaceSphere}, the VLM generated eight candidate
specifications:
\begin{align}
\phi_1:\quad
&\mathbf{F}_{[0,H_{\mathrm{task}}]}
\left(
\mathrm{Near}(e,o)
\land
\mathbf{F}_{[0,h_{\mathrm{reach}}]}
\mathrm{Grasped}(o)
\right),
\\
\phi_2:\quad
&\mathbf{F}_{[0,H_{\mathrm{task}}]}
\left(
\mathrm{Grasped}(o)
\land
\mathbf{F}_{[0,h_{\mathrm{displace}}]}
\left(
\mathrm{Displaced}(o)
\land
\mathbf{F}_{[0,h_{\mathrm{progress}}]}
\mathrm{Toward\_goal}(o)
\right)
\right),
\\
\phi_3:\quad
&\mathbf{F}_{[0,H_{\mathrm{task}}]}
\left(
\mathrm{Toward\_goal}(o)
\land
\mathbf{F}_{[0,h_{\mathrm{place}}]}
\left(
\mathrm{In\_receptacle}(o,r)
\land
\mathbf{F}_{[0,h_{\mathrm{release}}]}
\mathrm{Released}(o)
\right)
\right),
\\
\phi_4:\quad
&\mathbf{G}_{[0,H_{\mathrm{task}}]}
\left(
\mathrm{Released}(o)
\rightarrow
\mathbf{F}_{[0,h_{\mathrm{settle}}]}
\mathrm{Static}(o)
\right),
\\
\phi_5:\quad
&\mathbf{F}_{[0,H_{\mathrm{task}}]}
\left(
\mathbf{G}_{[0,h_{\mathrm{in\_receptacle}}]}
\mathrm{In\_receptacle}(o,r)
\right),
\\
\phi_6:\quad
&\mathbf{F}_{[0,H_{\mathrm{task}}]}
\left(
\mathbf{G}_{[0,h_{\mathrm{static}}]}
\mathrm{Static}(o)
\right),
\\
\phi_7:\quad
&\mathbf{F}_{[0,H_{\mathrm{task}}]}
\left(
\mathbf{G}_{[0,h_{\mathrm{in\_receptacle}}]}
\mathrm{In\_receptacle}(o,r)
\right)
\land
\mathbf{F}_{[0,H_{\mathrm{task}}]}
\left(
\mathbf{G}_{[0,h_{\mathrm{static}}]}
\mathrm{Static}(o)
\right),
\\
\phi_8:\quad
&\mathbf{F}_{[0,H_{\mathrm{task}}]}
\left(
\mathrm{Near}(o,r)
\land
\mathbf{F}_{[0,h_{\mathrm{enter}}]}
\left(
\mathrm{In\_receptacle}(o,r)
\land
\mathbf{F}_{[0,h_{\mathrm{release}}]}
\mathrm{Released}(o)
\right)
\right).
\end{align}

The candidate set represents the demonstrated sequence of approaching and
grasping the object, transporting it toward the receptacle, placing and
releasing it, and maintaining containment and stationarity after
placement.

\myparagraph{Example Grounding - PushCube:}

Let $p_o(t)$, $p_e(t)$, and $p_g$ denote the positions of the object, end-effector, and goal, respectively. We compute the end-effector distance $d_{eo}(t) = \|p_e(t)-p_o(t)\|_2$, planar goal distance $d_g(t) = \|p_{o,xy}(t)-p_{g,xy}\|_2$, displacement $d_{\mathrm{disp}}(t) = \|p_{o,xy}(t)-p_{o,xy}(0)\|_2$, cumulative goal progress $P(t) = d_g(0)-d_g(t)$, and instantaneous motion $m(t) = \|p_{o,xy}(t)-p_{o,xy}(t-1)\|_2$. The grounded atomic margins are then formulated as:
\begin{align}
\rho_{\mathrm{Near}}(t) &= \delta_{\mathrm{near}} - d_{eo}(t), 
& \rho_{\mathrm{TowardGoal}}(t) &= P(t) - \delta_{\mathrm{progress}}, \nonumber \\
\rho_{\mathrm{Displaced}}(t) &= d_{\mathrm{disp}}(t) - \delta_{\mathrm{displaced}}, 
& \rho_{\mathrm{AtGoal}}(t) &= \delta_{\mathrm{goal}} - d_g(t), \nonumber \\
\rho_{\mathrm{OnSupport}}(t) &= \delta_{z} - |z_o(t) - z_{\mathrm{support}}|, 
& \rho_{\mathrm{Static}}(t) &= \delta_{\mathrm{static}} - m(t), \label{eq:atomicmargins}
\end{align}
with $\rho_{\mathrm{Engaged}}(t) = \min\{\rho_{\mathrm{Near}}(t),\, m(t) - \delta_{\mathrm{engaged}}\}$.


The current PushCube parameters are
$\delta_{\mathrm{near}}, 
\delta_{\mathrm{engaged}},
\delta_{\mathrm{displaced}}, 
\delta_{\mathrm{progress}},
\delta_{\mathrm{goal}}, 
\delta_z,
\delta_{\mathrm{static}}$ and temporal bounds are
$ H_{\mathrm{task}},
h_{\mathrm{engage}},
h_{\mathrm{progress}},
h_{\mathrm{goal}},
h_{\mathrm{settle}}$.

\subsubsection{Hyperparameters}

For all four tasks, we use the standard ManiSkill PPO implementation without modifying the PPO optimization procedure. The common hyperparameters are: learning rate $3\times10^{-4}$, discount factor $\gamma=0.8$, GAE parameter $\lambda=0.9$, PPO clipping coefficient $0.2$, value-function coefficient $0.5$, target KL divergence $0.1$, $8$ PPO update epochs per batch, and $32$ minibatches. The actor and critic are three-layer MLPs with $256$ hidden units per layer and $\tanh$ activations.

For PushCube, we use $2048$ parallel environments, a rollout length of $20$, and a total training budget of $2$M environment steps.

For StackCube, we use $1024$ parallel environments, a rollout length of $50$, and a total training budget of $30$M environment steps. 

For LiftPegUpright, we use $1024$ parallel environments, a rollout length of $50$, and a total training budget of $25$M environment steps.

For PlaceSphere, we use $1024$ parallel environments, a rollout length of $50$, and a total training budget of $10$M environment steps.

\subsubsection{Source Videos \& VLM Prompts For Manipulation}
For each manipulation task, we record a single human demonstration. The demonstration is processed by the VLM using two sequential, task-agnostic prompts designed for general manipulation. The first prompt extracts a semantic event trace using the shared manipulation ontology, while the second maps this trace to a bank of parametric STL specifications. Importantly, the same prompts are applied to all manipulation tasks without any task-specific modification.

\subsubsubsection{Stage 1: Human Video to Semantic Manipulation Trace}

\begin{lstlisting}[style=promptstyle]
System Role:
Act as an expert Vision-Language Model specializing in Biomechanics, Robotic Manipulation, and Signal Temporal Logic (STL).

Context & Objective:
Analyze the provided video of a human performing a manipulation task. Your goal is to extract an embodiment-independent semantic description of the temporal manipulation structure. This analysis will be used to construct STL specifications for a robot with a different morphology, body dimensions, and controllers. Therefore, you must describe what happens semantically rather than reproducing the demonstrator's exact trajectory, human kinematics, speed, or incidental movements.

Ontology & Permitted Vocabulary:

Use only the following roles and predicates in formal semantic fields.

1. Entity Roles:

Effector: The component directly interacting with the environment (e.g., human hand, robot end-effector).
Object: A manipulable object whose state or position may change.
Receptacle: A container or region into which an object can be placed.
Handle: A graspable/interactable component used to manipulate an articulated object.
Articulated_part: A movable articulated component (e.g., drawer, door).
Goal: A visually identifiable destination or goal region.
Support: A surface or structure physically supporting an object.

2. States & Events (Predicates):

Near(a, b): Entity a is spatially close to entity b.
Engaged(effector, x): The effector is actively interacting with x to change its state/motion.
Grasped(object): The object is securely held by the effector.
Released(object): The object is no longer grasped by the effector.
Displaced(object): The object has moved meaningfully from its initial location.
Toward_goal(object): The object has made meaningful progress toward a visual goal.
Upright(object): The object is in an approximately upright or standing orientation.
At_goal(object): The object occupies a visually identifiable goal region.
On_support(object, support): The object remains supported by the intended supporting surface.
In_receptacle(object, receptacle): The object occupies the interior/accepted region of a receptacle.
Static(x): Entity x is approximately stationary.
Open(articulated_part): The articulated part is in an open configuration.

3. Temporal Relations:

before
after
overlaps
persists_until

4. Observation Classes:

Essential_candidate: Visually supported, embodiment-independent temporal relationships critical to describing the manipulation.
Terminal_candidate: An observed state/event that plausibly represents completion or the resulting outcome.
Incidental: Visually present but unnecessary for formal description.

Strict Constraints & Rules:

1. A role, state, or event may be absent from the video.
   Do not force every role or state to be populated.

2. Do NOT infer joint angles, velocities, or numerical thresholds.

3. Do NOT use human-specific variables
   (e.g., hand_x, wrist_angle, elbow_pose)
   inside semantic predicates.

4. Do NOT interpret exact human timing for the future robot.

Task Requirements:

1. Identify the visually relevant entities in the video and map them to the permitted semantic roles.

2. Formulate the temporal relationships among observed events
   using ONLY the permitted predicates.

3. Identify persistent properties only when the video gives clear evidence that a property is maintained for a meaningful interval.

4. Note any critical ambiguities or visual occlusions.

Output Format:

Return valid JSON ONLY. Do not include markdown formatting,
code blocks, or conversational text outside the JSON structure.
Use exactly this schema:

{
  "manipulation_context": {
    "confidence": 0.0,
    "evidence":
      "Brief visual justification of the overall task being performed"
  },

  "entity_mapping": {
    "effector":
      "Description of what serves as the effector (or null)",
    "object":
      "Description of the object (or null)",
    "receptacle":
      "Description of the receptacle (or null)",
    "handle":
      "Description of the handle (or null)",
    "articulated_part":
      "Description of the articulated part (or null)",
    "goal":
      "Description of the goal region (or null)",
    "support":
      "Description of the support surface (or null)"
  },

  "visibility": {
    "effector": "clear | partial | unclear | n/a",
    "object": "clear | partial | unclear | n/a",
    "receptacle": "clear | partial | unclear | n/a",
    "handle": "clear | partial | unclear | n/a",
    "articulated_part": "clear | partial | unclear | n/a",
    "goal": "clear | partial | unclear | n/a",
    "support": "clear | partial | unclear | n/a"
  },

  "observed_events": [
    {
      "event_id": "e1",
      "state_or_event": "e.g., Grasped(object)",
      "event_class":
        "essential_candidate | terminal_candidate | incidental",
      "confidence": 0.0,
      "visual_evidence":
        "Brief explanation of supporting visual evidence"
    }
  ],

  "temporal_relations": [
    {
      "event_1": "e1",
      "relation": "before | after | overlaps | persists_until",
      "event_2": "e2",
      "confidence": 0.0
    }
  ],

  "persistent_properties": [
    {
      "state_or_event":
        "e.g., On_support(object, support)",
      "confidence": 0.0,
      "visual_evidence": "Brief explanation"
    }
  ],

  "missing_concepts": [
    {
      "description":
        "Important observed concept not expressible by the strict ontology",
      "importance": "low | medium | high"
    }
  ],

  "uncertainties": [
    "Brief statement describing an important ambiguity or occlusion"
  ]
}
\end{lstlisting}

\subsubsubsection{Stage 2: Semantic Manipulation Trace to PSTL Specifications}

\begin{lstlisting}[style=promptstyle]
System Role:
Act as an expert Formal Methods Engineer specializing in Logic Synthesis for Robotic Control.

Context & Objective:
You are constructing Parametric Signal Temporal Logic (PSTL)
specifications based on a semantic manipulation trace.
This trace (provided as a JSON input) was extracted from a human demonstration video.

Your objective is to translate this trace into embodiment-independent, candidate temporal specifications that describe the demonstrated behavior. These specifications will eventually be grounded to a robot with a different morphology to generate a dense, continuous reward signal for reinforcement learning.

Strict Constraints:

- Trace-Dependent Reasoning: You must reason only from the supplied semantic trace.
- No Invention: Do not invent contact events, predicates, or temporal relations that are absent from the input data. Use a predicate only if it is explicitly supported by the trace.

Symbolic Grounding:

All temporal constants and numerical temporal parameters must remain purely symbolic (e.g., H_task, h_persist, h_reach) for later numerical grounding during training.

Allowed PSTL Grammar:

You are restricted to the following exact textual grammar.
(Note: mu represents an atomic predicate; phi represents a sub-formula; H and h represent symbolic temporal bounds).

Atomic predicate:
mu

Conjunction:
phi_1 AND phi_2

Bounded eventuality:
F_[0,H](mu)

Bounded persistence:
G_[0,H](mu)

Response requirement:
G_[0,H](mu_1 -> F_[0,h](mu_2))

Two-event ordered sequence:
F_[0,H](mu_1 AND F_[0,h1](mu_2))

Three-event ordered sequence:
F_[0,H](mu_1 AND F_[0,h1](mu_2 AND F_[0,h2](mu_3)))

Terminal stability:
F_[0,H](G_[0,h](mu))

Conjunctions of independently valid formulas are permitted.

Output Format:

Return valid JSON only, using exactly the structure below.
Do not include markdown code blocks, conversational text, or
explanations outside the JSON.

{
  "ontology_sufficiency":
    "sufficient | partially_sufficient | insufficient",

  "ontology_note":
    "Brief explanation of whether the available grammar and predicates adequately captured the trace.",

  "candidate_specifications": [
    {
      "candidate_id": "phi_1",

      "description":
        "Plain English description of the behavior this formula enforces",

      "formula":
        "PSTL formula using exactly the allowed textual grammar",

      "predicates_used": [
        "instantiated predicate (e.g., Grasped(object))"
      ],

      "supporting_event_ids": [
        "e1",
        "e2"
      ],

      "excluded_incidental_event_ids": [
        "e4"
      ],

      "ordered_events": [
        "instantiated predicate 1",
        "instantiated predicate 2"
      ],

      "persistent_requirements": [],

      "symbolic_parameters": [
        "H_task",
        "h1"
      ],

      "semantic_complexity":
        "low | moderate | high",

      "rationale":
        "Brief explanation grounded only in the provided semantic trace"
    }
  ],

  "stage1_ambiguities_affecting_specification": [
    "Brief explanation of any ambiguity in the input trace that impacted formula construction"
  ]
}
\end{lstlisting}
\end{document}

%% file: math_commands.tex
\usepackage{amsmath,amsfonts,bm}

\def\eqref#1{equation~\ref{#1}}

\def\1{\bm{1}}

\DeclareMathAlphabet{\mathsfit}{\encodingdefault}{\sfdefault}{m}{sl}
\SetMathAlphabet{\mathsfit}{bold}{\encodingdefault}{\sfdefault}{bx}{n}



%% file: inc-packages.tex
\usepackage{color,xcolor}
\usepackage{epsfig}
\usepackage{graphicx}

\usepackage[export]{adjustbox}
\usepackage{array}
\usepackage{booktabs}
\usepackage{colortbl}
\usepackage{wrapfig}
\usepackage{hhline}
\usepackage{multirow}
\usepackage{subcaption} 
\usepackage{wrapfig}
\usepackage{floatflt}
\usepackage{siunitx}
\usepackage{caption}

\usepackage{amsmath,amsfonts,amssymb,amsthm}
\usepackage{mathtools}  
\usepackage{bm}
\usepackage{nicefrac}
\usepackage{microtype}
\usepackage[T1]{fontenc}
\usepackage[ansinew]{inputenc}

\usepackage{changepage}
\usepackage{extramarks}
\usepackage{fancyhdr}
\usepackage{lastpage}
\usepackage{setspace}
\usepackage{soul}
\usepackage{xspace}

\usepackage{url}

\usepackage{enumerate}
\usepackage{todonotes} 
\usepackage{enumitem}  

\usepackage{titlesec}

\usepackage{makecell}

\usepackage{algorithm}
\usepackage[noend]{algpseudocode}
\renewcommand{\algorithmicrequire}{\textbf{Input:}}
\renewcommand{\algorithmicensure}{\textbf{Output:}}
\usepackage{framed}
\definecolor{shadecolor}{gray}{0.95}

\usepackage{xparse}
\usepackage{etoolbox}


%% file: inc-macros.tex
\DeclareMathAlphabet{\mathbbb}{U}{bbold}{m}{n}

\newcolumntype{L}[1]{>{\raggedright\let\newline\\\arraybackslash\hspace{0pt}}m{#1}}
\newcolumntype{C}[1]{>{\centering\let\newline\\\arraybackslash\hspace{0pt}}m{#1}}
\newcolumntype{R}[1]{>{\raggedleft\let\newline\\\arraybackslash\hspace{0pt}}m{#1}}

\usepackage{pifont}
\newcommand{\ignore}[1]{}

\makeatletter
\DeclareRobustCommand\onedot{\futurelet\@let@token\@onedot}
\def\@onedot{\ifx\@let@token.\else.\null\fi\xspace}

\makeatother

\definecolor{RayColor}{rgb}{0,0.08,1}
\definecolor{URL}{HTML}{0000EE}
\definecolor{MyDarkGreen}{rgb}{0.02,0.6,0.02}
\definecolor{MyDarkRed}{rgb}{0.8,0.02,0.02}
\definecolor{MyDarkOrange}{rgb}{0.40,0.2,0.02}
\definecolor{MyPurple}{RGB}{111,0,255}
\definecolor{MyRed}{rgb}{1.0,0.0,0.0}
\definecolor{MyGold}{rgb}{0.75,0.6,0.12}
\definecolor{MyDarkgray}{rgb}{0.66, 0.66, 0.66}

\newcommand{\myparagraph}[1]{\noindent\textbf{#1}}

\theoremstyle{plain}

\usepackage{titlesec}
\titlespacing*{\section}{0pt}{3pt plus 2pt minus 2pt}{3pt plus 2pt minus 2pt}
\titlespacing*{\subsection}{0pt}{2pt plus 2pt minus 2pt}{2pt plus 2pt minus 2pt}
\titlespacing*{\subsubsection}{0pt}{1pt plus 1pt minus 1pt}{1pt plus 1pt minus 1pt}

\definecolor{promptshade}{RGB}{246,248,251}
\newsavebox{\promptboxsave}
  {\vspace{2pt}\end{minipage}%
  \end{lrbox}%
  \colorbox{promptshade}{\usebox{\promptboxsave}}\par\medskip}